\documentclass[runningheads]{llncs}

\usepackage{eccv}

\usepackage{eccvabbrv}

\usepackage{graphicx}
\usepackage{booktabs}
\usepackage{fontawesome5}
\usepackage{pifont}
\newcommand{\cmark}{\ding{51}}
\newcommand{\xmark}{\ding{55}}
\usepackage{tikz}
\usetikzlibrary{positioning,fit,backgrounds,arrows.meta}
\usepackage{float}
\usepackage{array}

\usepackage{placeins}    
\usepackage{listings}

\usepackage[accsupp]{axessibility}  

\usepackage{hyperref}

\usepackage{orcidlink}

\definecolor{slotcol}{HTML}{B23B3B}
\definecolor{casecol}{HTML}{1A5276}
\definecolor{dropcol}{HTML}{808080}

\begin{document}

\title{Dissecting Agentic Forensics: The Role of Triage, Prompting, and Evidence Arbitration in Open-World Fake Image Detection}

\titlerunning{Dissecting Agentic Forensics}

\author{Xianlong Li\inst{2} \and Pietro Bongini\inst{1} \and Niccol{\'o} Pancino\inst{1} \and Marco Blanchini\inst{2} \and Benedetta Tondi\inst{1} \and Mauro Barni\inst{1}}
\authorrunning{Li et al.}
\institute{University of Siena, Italy \and IMT School for Advanced Studies Lucca, Italy}

\maketitle

\begin{abstract}
Image forensics is increasingly an open-world problem: manipulations range from fully synthetic images to localized edits, splicing and swapping, while most forensic detectors remain specialized to a single manipulation family.
Agentic AI has recently emerged as a promising solution. In principle, such systems can assess the reliability of individual detectors, identify out-of-scope evidence, and arbitrate conflicting reports. However, it remains unclear which components actually drive performance and whether their benefits persist under distribution shift. To answer these questions, we study a training-free agentic framework built around specialist detectors, per-detector triage, and conflict-aware evidence arbitration. Using six configurations and three multimodal large language model backbones, we dissect the role of triage, prompting, and reasoning quality on both in-distribution and out-of-distribution data.
Our results show that naive detector fusion suffers from severe false-positive rates on authentic images. Triage and prompting consistently improve performance by filtering unreliable evidence and exposing detector limitations. However, the dominant factor is represented by reasoning itself: A stronger judge substantially outperforms a weaker one, particularly under distribution shift. Most notably, manipulation recall is nearly saturated across all configurations, indicating that the main challenge of open-world image forensics is not detecting manipulations, but calibrating trust in specialized forensic tools and arbitrating conflicting evidence.

  \keywords{Image Forensics \and Agentic AI \and Multi-Agent Systems}
\end{abstract}

\vspace{0.5em}
\noindent\rule{\textwidth}{0.4pt}
\vspace{0.2em}

\noindent{\footnotesize
Accepted at the 2026 Workshop on AI for Multimedia Forensics and Disinformation
Detection (AI4MFDD), ECCV 2026. This is the authors' version of the contribution.
\par}

\vspace{0.2em}
\noindent\rule{\textwidth}{0.4pt}

\section{Introduction}
\label{sec:intro}

Advances in generative modeling have made image manipulation more convincing, cheaper, and more pervasive, while broadening the forms it takes: an image may be synthesized in its entirety, edited locally by diffusion-based methods~\cite{rombach2022ldm} or generative inpainting, assembled by splicing content from other sources, or altered by face swapping, among other techniques.
Determining whether, and how, an image has been manipulated is therefore central to media authenticity and information integrity~\cite{verdoliva2020media}.
The difficulty is that image forensics is inherently an open-world problem: in realistic scenarios, a forensic system must accept an arbitrary image and remain reliable whichever of these manipulations, if any, it has undergone.

Most forensic detectors are specialists: each targets one manipulation family and is accurate within it, but its behaviour outside that family is neither defined nor dependable, and such detectors are known not to generalise across generators or manipulation families~\cite{ojha2023universal,wang2020cnn}.
The natural response - to run several specialists and fuse their outputs, an idea with a long history in forensic decision fusion~\cite{fontani2013framework} - breaks down precisely when they disagree.
A detector applied to content it was never designed to judge may report a manipulation with high confidence, yielding a false positive on an authentic image, or stay silent on a manipulation beyond its scope, yielding a false negative.
A flat aggregation of scores has no principled basis for deciding which detector to believe. Resolving such conflicts is therefore a matter of judgement rather than arithmetic: it requires assessing whether a detector is applicable to the image at hand and weighing its evidence against competing reports.

Agentic approaches have recently applied multimodal large language models (MLLMs) to this judgement~\cite{liang2025aifo,zhang2025forenagent}. However, they are typically developed and evaluated as monolithic pipelines. As a result, it remains unclear which components drive performance, how much of the observed gains stem from the underlying model rather than the pipeline design, and whether these gains persist under distribution shift. These questions are the subject of this paper.

To study these questions, we adopt a deliberately simple agentic framework whose components can be isolated and evaluated independently. The design is guided by two observations. First, specialist detectors are frequently applied outside their intended scope, suggesting the need for an explicit triage stage that assesses whether a detector's output should be considered evidence at all. Second, detectors may produce false positives, false negatives, and mutually conflicting reports, suggesting the need for a judge that reasons explicitly about detector limitations rather than blindly aggregating their outputs.

The resulting framework requires no training of its own. A fixed pool of pre-trained specialist detectors is applied to an input image; a per-detector auditor agent triages each output, deciding whether it should be admitted as evidence; and a conflict-aware judge arbitrates the retained evidence into a final verdict—a binary decision together with an attribution of the evidence on which that decision rests (\cref{fig:framework}).

We instantiate the framework with seven publicly available detectors spanning five manipulation families. The detector pool is extensible: supporting a new manipulation family requires only registering an additional specialist detector, with no retraining.
We then use this framework as an experimental testbed to study three aspects of agentic forensic reasoning: evidence triage, explicit awareness of detector limitations, and conflict-aware arbitration. Through a controlled set of ablations, we isolate the contribution of each component and evaluate their interaction with different MLLM backbones.
Because open-world forensic systems must operate on previously unseen content, all experiments are performed both in-distribution and out-of-distribution. This allows us to assess not only overall performance, but also the robustness of the different reasoning components under distribution shift.

Our experiments reveal that the benefits of agentic forensics do not stem equally from all components. Evidence triage and explicit awareness of detector limitations provide consistent improvements by filtering unreliable evidence and exposing detector failure modes. However, these benefits can only be fully exploited by sufficiently capable reasoning models, whose quality ultimately dominates performance. We further find that manipulation recall is already close to saturation, and that performance differences arise primarily from the ability to reject false alarms on authentic images. These findings suggest that the central challenge of open-world image forensics is not manipulation detection per se, but the calibration of trust in specialized forensic tools.

With the above ideas in mind, the contributions of the paper can be summarised as follows:
\begin{itemize}
\item A simple, training-free agentic framework for open-world image forensics, designed as an experimental testbed for combining specialist detectors through evidence triage and arbitration.
\item A controlled dissection of the role of triage, prompting, and evidence arbitration across multiple MLLM backbones under both in-distribution and out-of-distribution conditions.
\item An empirical study of the factors driving performance in agentic image forensics.
\item A simple and measurable form of evidence attribution, identifying the detector deemed decisive by the arbitration process in addition to the final binary verdict.
\end{itemize}

\section{Related Work}
\label{sec:related}

Our work builds on three principal lines of research, which we review in turn: forensic detectors specialised to a single manipulation family, approaches that jointly address multiple manipulation families, and the emerging integration of agentic reasoning with forensic tools, to which our method belongs.

\subsubsection{Single-Task Forensic Detection.}
The majority of forensic detectors are designed for a single manipulation family~\cite{verdoliva2020media}.
Whole-image generation detectors determine whether an image is entirely synthetic, relying on generator-specific fingerprints~\cite{wang2020cnn}, CLIP-based representations~\cite{cozzolino2024clip,ojha2023universal}, cross-generator inconsistencies~\cite{yang2025d3}, or feature-subspace decompositions~\cite{yan2025effort}.
Face-manipulation detectors concentrate on a single tampered face and its associated blending artefacts, and are typically evaluated on the FaceForensics++ benchmark~\cite{rossler2019faceforensics}.
Localised-edit detectors, in contrast, produce pixel-level tampering maps for operations such as splicing, copy-move, and inpainting~\cite{guillaro2023trufor}, with recent methods further specialising in diffusion-based inpainting~\cite{costanzino2025radar,cai2026nfavit}.
Each detector is thus calibrated to a specific manipulation family and data distribution, and its performance tends to deteriorate when applied to images subjected to other manipulations.

\subsubsection{Multi-Task Forensic Detection.}
Since the manipulation family is typically unknown \emph{a priori}, a second major research direction addresses multiple manipulation families simultaneously.
Classical approaches perform decision-level fusion over a set of specialised forensic tools, combining their outputs into a single verdict and weighing each according to its estimated reliability under the observed acquisition and processing conditions~\cite{fontani2013watchful}; decision fusion based on Dempster--Shafer evidence theory constitutes a canonical formulation of this paradigm~\cite{fontani2013framework}.
Deep learning approaches instead train a single network to jointly detect and localise multiple manipulation families without family-specific supervision~\cite{wu2019mantranet,dong2023mvssnet}.
More recent work extends this with multimodal architectures that additionally produce a natural-language rationale~\cite{xu2025fakeshield,li2024forgerygpt}.
Although these approaches increase the coverage of manipulation families, the fusion rules in classical systems are manually specified, and the learned multimodal models are trained on, and therefore constrained by, a fixed and predefined manipulation taxonomy.

\subsubsection{Agentic Forensics.}
A third, more recent line of research endows the model with agency: rather than emitting a verdict in a single forward pass, the model iteratively reasons and invokes external tools~\cite{schick2023toolformer}.
This direction builds on agentic reasoning frameworks such as ReAct, which interleave explicit reasoning steps with tool calls~\cite{yao2023react}, as well as on methods that enhance the robustness and reliability of tool usage via self-critique and inference-time feedback mechanisms~\cite{gou2024critic,ta2026reinforced}.
In the forensic domain, AIFo casts AI-generated image detection as a multi-agent reasoning problem over provenance information, metadata, classifier outputs, and vision--language cues~\cite{liang2025aifo}; EvoGuard trains an agent via reinforcement learning to dynamically select among a set of off-the-shelf detectors~\cite{zhu2026evoguard}; and ForenAgent employs a multimodal model that iteratively invokes hand-crafted low-level forensic cues under reinforcement fine-tuning~\cite{zhang2025forenagent}.
These systems are most closely related to our work, but they are all designed to answer a single binary decision problem---authentic versus manipulated---for a specific manipulation task, rather than to identify and localise the manipulation family in an open-world setting. They rely on task-specific training or explicitly engineered low-level cues.
By contrast, our framework orchestrates a pool of pre-trained, state-of-the-art single-task detectors spanning multiple manipulation families without any additional training, identifies the manipulation's family and region rather than only its presence, and treats the scope conflicts and out-of-competence errors that combining specialists entails as first-class problems.

\section{Method}
\label{sec:method}

\subsection{Problem Setting and Overview}
\label{sec:method-setting}

We address open-world image forensics.
Given an image $x$ of unknown provenance, the system returns a verdict $(y, m)$: a binary authenticity label $y \in \{0, 1\}$ (authentic or manipulated) and, when $y = 1$, the manipulation family $m \in \mathcal{M}$ responsible for the manipulation.
We place no prior on the family, and we assume that a manipulated image belongs to a single family of $\mathcal{M}$. Composite manipulations are beyond the scope of this paper.

Covering $\mathcal{M}$ requires a pool of \emph{targeted} detectors, each specialised to one manipulation family.
Combining them raises the two problems set out in \cref{sec:intro}, both absent in the single-family case.
First, a detector may be applied \emph{outside its domain of competence}: when confronted with content that its design does not encompass, it may return an uninformative score, either producing a false positive on an authentic image or failing to flag a genuine manipulation.
Second, because at most one family can correctly account for $x$, detectors targeting different manipulation families may yield \emph{conflicting evidence}: several detectors can simultaneously report a manipulation even though they cannot all be correct.

These two problems motivate the two principal components around which our framework is organised.
\emph{Triage} addresses the first: before a detector's output is admitted as evidence, an agent first assesses whether that detector is applicable to the image under analysis.
\emph{Arbitration} addresses the second: a single agent aggregates the admitted, potentially conflicting reports and decides which of them, if any, to regard as reliable.
Orthogonal to both is a third, equally important design dimension, \emph{prompting}: what the agents are explicitly told about each detector's domain of competence and about the possibility of conflict, rather than leaving such information implicit.
We treat triage, prompting, and arbitration as the three central design choices to be analysed empirically (\cref{sec:eval}).

\begin{figure}[t]
  \centering
  \resizebox{\textwidth}{!}{%
  \begin{tikzpicture}[
      font=\small, >=Latex,
      det/.style={draw=black!35, rounded corners=1pt, fill=white, align=left,
                  inner sep=3pt, text width=44mm, minimum height=5mm, font=\footnotesize},
      grp/.style={font=\bfseries\footnotesize, inner sep=1pt, anchor=west},
      aud/.style={draw=green!50!black!55, rounded corners=2pt, fill=green!14,
                  minimum width=9mm, minimum height=4.5mm, inner sep=1pt, align=center, font=\footnotesize},
      box/.style={draw, rounded corners=2pt, align=center, inner sep=5pt},
      code/.style={box, fill=blue!4, draw=blue!40},
      agent/.style={box, fill=green!12, draw=green!50!black!55},
      arr/.style={->, semithick, draw=black!60},
    ]
    \node[grp] (g1) {Whole-image generation};
    \node[det, below=1mm of g1.south west, anchor=north west] (d1) {Cozzolino \etal~\cite{cozzolino2024clip}};
    \node[det, below=0.7mm of d1.south west, anchor=north west] (d2) {D\textsuperscript{3}~\cite{yang2025d3}};
    \node[grp, below=1.8mm of d2.south west, anchor=north west] (g2) {Face manipulation};
    \node[det, below=1mm of g2.south west, anchor=north west] (d3) {Effort~\cite{yan2025effort}};
    \node[det, below=0.7mm of d3.south west, anchor=north west] (d4) {ForensicsAdapter~\cite{cui2025forensicsadapter}};
    \node[grp, below=1.8mm of d4.south west, anchor=north west] (g3) {Local edits};
    \node[det, below=1mm of g3.south west, anchor=north west] (d5)
      {RADAR~\cite{costanzino2025radar}\\{\scriptsize\textcolor{black!55}{diffusion inpainting}}};
    \node[det, below=0.7mm of d5.south west, anchor=north west] (d6)
      {NFA-ViT~\cite{cai2026nfavit}\\{\scriptsize\textcolor{black!55}{diffusion\,/\,GAN inpainting}}};
    \node[det, below=0.7mm of d6.south west, anchor=north west] (d7)
      {TruFor~\cite{guillaro2023trufor}\\{\scriptsize\textcolor{black!55}{inpainting, splicing\,/\,copy-move}}};
    \begin{scope}[on background layer]
      \node[code, fit=(g1)(d1)(d2)(g2)(d3)(d4)(g3)(d5)(d6)(d7), inner sep=8pt,
            label={[font=\bfseries]above:Stage 0: detector pool}] (s0) {};
    \end{scope}

    \node[code, left=14mm of s0, text width=22mm] (img)
      {Input image $x$\\[2pt]{\scriptsize of unknown provenance}};
    \draw[arr] (img.east) -- node[above,font=\scriptsize]{run on all} (s0.west);

    \foreach \i in {1,...,7}{
      \node[aud, right=22mm of d\i.east] (a\i) {\faRobot};
      \draw[arr] (d\i.east) -- (a\i.west);
    }
    \begin{scope}[on background layer]
      \node[agent, fit=(a1)(a7), inner sep=6pt,
            label={[font=\bfseries,align=center]above:Stage~1\\{\normalfont\footnotesize per-detector triage}}] (s1) {};
    \end{scope}

    \node[agent, right=18mm of s1.east, text width=24mm] (judge)
      {{\faRobot}\\[2pt]\textbf{Stage 2}\\{\footnotesize conflict-aware judge}};
    \draw[arr] (s1.east) -- node[above,font=\scriptsize]{retained} node[below,font=\scriptsize]{only} (judge.west);

    \node[code, align=left, text width=46mm, right=13mm of judge.east] (v)
      {\textbf{Verdict}\\[3pt]\footnotesize
       $y$: authentic or manipulated\\[1.5pt]
       $m$: manipulation family\\[1.5pt]
       decisive detector};
    \draw[arr] (judge.east) -- (v.west);
  \end{tikzpicture}}
  \caption{Our framework as a three-stage pipeline.
    \textbf{Stage~0} runs the full pool of pre-trained detectors on the input image, grouped by manipulation family: whole-image generation, face manipulation, and local edits, with each local-edit detector annotated by the manipulations it targets.
    \textbf{Stage~1} audits each detector with its own agent (one auditor per detector), which keeps or discards that detector's output; only the retained detectors are passed on.
    \textbf{Stage~2} arbitrates them with a single conflict-aware judge, returning the verdict $(y, m)$, where $m$ is the family of the decisive detector.
    Splicing and copy-move are treated as a single family because TruFor is the only pooled detector covering them; the localisation detectors (RADAR, NFA-ViT, TruFor) additionally return a tampering map.
    Blue: automated steps; green: MLLM agents.}
  \label{fig:framework}
\end{figure}
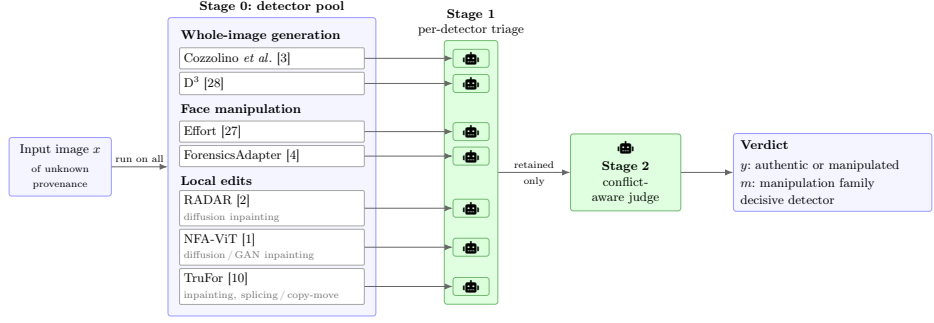

\Cref{fig:framework} shows the pipeline of our agentic framework, which requires no training of its own and proceeds in three stages.
In Stage~0, the full pool of detectors is run on $x$; each returns a manipulation score, and the localisation detectors additionally return a tampering map.
In Stage~1, a dedicated \emph{auditor} agent triages each detector independently, discarding any whose output is not credible for the image at hand: for instance, a face manipulation detector applied to an image with no visible face, or a tampering map scattered over natural texture rather than concentrated on a plausible edit.
In Stage~2, a single \emph{judge} agent integrates the output of the retained detectors, resolves possible conflicts, and emits the verdict $(y, m)$ including the indication of the detector whose evidence it deems decisive.
The agents are MLLMs prompted to read the available evidence and produce these judgements.
We describe the detector pool in \cref{sec:method-pool} and the agents in \cref{sec:method-agents}.

\begin{table}[t]
  \centering
  \caption{The seven pre-trained detectors and their declared scope (the manipulation family or families each targets). Following our setting, splicing and copy-move form a single family, since TruFor is the only pooled detector covering them. Localisation detectors additionally produce a dense tampering map.}
  \label{tab:pool}
  \small
  \begin{tabular}{@{}lll@{}}
    \toprule
    Detector & Declared scope & Output \\
    \midrule
    Cozzolino \etal~\cite{cozzolino2024clip} & Whole-image generation & score \\
    D\textsuperscript{3}~\cite{yang2025d3} & Whole-image generation & score \\
    Effort~\cite{yan2025effort} & Face manipulation & score \\
    ForensicsAdapter~\cite{cui2025forensicsadapter} & Face manipulation & score \\
    RADAR~\cite{costanzino2025radar} & Diffusion inpainting & score + map \\
    NFA-ViT~\cite{cai2026nfavit} & Diffusion / GAN inpainting & score + map \\
    TruFor~\cite{guillaro2023trufor} & Diffusion / GAN inpainting, splicing/copy-move & score + map \\
    \bottomrule
  \end{tabular}
\end{table}

\subsection{Detector Pool and Tool Cards}
\label{sec:method-pool}

We instantiate the pool of detectors with seven publicly available methods whose combined competence spans the five families of $\mathcal{M}$: whole-image generation, face manipulation, diffusion inpainting, GAN inpainting, and splicing/copy-move (\cref{tab:pool}).
Each detector is treated as a black box that emits a scalar indication of manipulation and, for the three localisation detectors, a dense tampering map.

\begin{table}[t]
  \centering
  \caption{The tool card of TruFor, shown to make concrete what every card declares: the detector's scope and the attack classes it covers, its decision threshold, how to read its score, and how to read its heatmap. All seven cards share this structure.}
  \label{tab:toolcard}
  \footnotesize
  \begin{tabular}{@{}l >{\raggedright\arraybackslash}p{0.72\textwidth}@{}}
    \toprule
    Field & Content (TruFor) \\
    \midrule
    \texttt{scope\_in} & Local-manipulation localisation: splicing, copy-move, and inpainting (including diffusion-based); produces an image-level score and a pixel-level anomaly map. \\
    \texttt{attack\_classes} & \texttt{inpaint\_diffusion}, \texttt{inpaint\_gan\_traditional}, \texttt{splice\_copymove} \\
    \texttt{threshold} & \texttt{integrity\_score} in $[0,1]$; decision boundary at $0.5$ (higher means more likely tampered). \\
    \texttt{signal\_interp.} & A higher score means more likely tampered; a low score does not by itself rule out manipulation if the anomaly map shows a concentrated hot region at a plausible tamper location. \\
    \texttt{heatmap\_desc} & Two-panel figure: left, the original with the anomaly map overlaid; right, a pure jet heatmap (red = likely manipulated, blue = pristine). \\
    \bottomrule
  \end{tabular}
\end{table}

Crucially, each detector is registered together with a \emph{tool card}: a short, externally authored description that is the agents' only knowledge of the detector.
A card declares the detector's \emph{scope} (the manipulation families it targets), its decision threshold (fixed at $0.5$ for every detector in our pool), and how to read its output: the score and, for the localisation detectors, the tampering heatmap (\cref{tab:toolcard}).
It is therefore the card that lets an agent recognise a score as out of scope, and so uninformative, rather than as evidence.
Exposing each detector's competence in this way is a deliberate design choice rather than an implementation detail, and its effect is one of the objective of our study (\cref{sec:eval}).
\Cref{tab:toolcard} shows one card in full; the remaining six are given as supplementary material.

The detectors' native outputs are heterogeneous and not directly comparable, so a per-detector calibration maps each score onto a common range $s \in [0, 1]$, with a shared threshold equal to $0.5$ separating a detector that reports a manipulation from a silent one.
Because a detector's family membership and output semantics reside entirely in its tool card rather than in the surrounding code, extending the pool to a new family reduces to registering one further detector and its card.

\subsection{Agentic Triage and Arbitration}
\label{sec:method-agents}

\paragraph{Triage: Per-Detector Auditing.}
Stage~1 addresses the first problem outlined before: out-of-competence detection.
For every detector that produced an output, a dedicated auditor agent decides whether that output is credible evidence on the image at hand, and excludes it from all downstream processing if it is not.
The audit turns on two questions: whether the image actually contains the kind of content the detector targets (a scope check, taken as a hard prerequisite), and, if so, whether the detector's signal is internally coherent, meaning a score clearly separated from the common $0.5$ decision threshold  (\cref{sec:method-pool}): a score resting on this boundary leaves the detection undecided and hence uninformative, and, for localisation detectors, a map concentrated on a plausible edit rather than dispersed over natural texture.
We assign one narrowly scoped agent to each detector, rather than reading the detector's scalar directly, because deciding whether that scalar is credible \emph{for this particular image} demands contextual judgement the detector cannot provide.
Concretely, each auditor is an MLLM that reads the image, the detector's score and map, and its tool card, and returns a verdict (keep or discard) together with a brief written justification.
Assigning one agent per detector follows a standard multi-agent design, in which a task is decomposed into narrowly scoped, single-purpose agents rather than handled by one monolithic agent~\cite{wu2023autogen,hong2024metagpt}.
The opposite choice, folding the audit into the judge so that a single agent both audits and arbitrates, is one of the comparison configurations consider in our experiments (\cref{sec:eval}).

\paragraph{Arbitration: The Conflict-Aware Judge.}
Stage~2 addresses the second problem: conflicting evidence.
A single judge agent receives, for each retained detector, its score, the output map (if any), and a tool card together with that detector's auditor verdict and justification. The judge also has access to the analyzed image itself. From these data, the judge produces the verdict $(y, m)$.
Because at most one family can explain the image, detectors of differing scope that simultaneously report a manipulation cannot all be correct; the judge must decide which one, if any, reflects a genuine manipulation.
That detector is the framework's \emph{evidence attribution}: the family $m$ is read directly off its declared scope, so naming the decisive detector and naming $m$ are equivalent operations.
We make this attribution a primary output, not a by-product of the binary decision.
A detector applied beyond its competence frequently reports a manipulation on out-of-distribution content, so a framework can return the correct binary label while crediting the wrong detector - a verdict that is right only by accident and whose underlying explanation is unsound.
For this reason, in \cref{sec:eval}, we also verify that the decisive detector be appropriate for the image at hand, in addition to checking if $y$ is correct or not.

The manner in which the judge is prompted is fundamental to its behaviour. The prompt does not request a verdict directly; instead, it specifies an explicit, stepwise protocol: first, inspect the image directly; next, enumerate the cross-scope conflicts among detectors that report a manipulation; then, evaluate each report in light of both the image content and the detector’s stated domain of competence; and only thereafter output a judgment.
The judge’s reasoning relies critically on two forms of guidance encoded in this prompt.
The first concerns conflict: since at most one family can explain the image, retained detectors whose declared scopes differ cannot jointly detect a true manipulation, and the prompt instructs the judge to identify such cross-scope disagreements and resolve them rather than average or otherwise smooth them away.
The second concerns reliability: a detector remains an imperfect classifier, so a positive report may reflect either a genuine detection or a false alarm. Examples include near-threshold scores, tampering maps that highlight natural texture rather than plausible edits, or isolated positive reports unsupported by other detectors targeting the same manipulation family. The prompt therefore instructs the judge to reconcile each detector report with its own interpretation of the image, weighing the credibility of the evidence before reaching a final decision.

Both are deliberate prompting design choices; removing them, so that the judge aggregates the same retained detector outputs without explicit awareness of conflicts or of residual false-positive risk, is the basis for the prompting ablation study described in \cref{sec:eval}.
The complete protocol and all prompts are provided verbatim in the supplementary material.

\section{Evaluation Design and Results}
\label{sec:eval}

\subsection{Datasets}
\label{sec:eval-data}
All configurations (\cref{sec:eval-variants}) are evaluated on two data collections, each covering the five manipulation families forming $\mathcal{M}$ alongside authentic imagery.
Each collection is drawn from a larger annotated pool (7,000 images for the in-distribution collection, 7,496 for the out-of-distribution), by including a fixed number of images per manipulation family.
In this way, we built an in-distribution evaluation set of $240$ images ($135$ manipulated, $105$ authentic) and an out-of-distribution set of $200$ images ($100$ manipulated - $20$ per family - and $100$ authentic).
The \emph{in-distribution} collection is organised by detector: for each of the seven pooled detectors, we sample authentic images and images of the manipulation family targeted by that detector from the benchmark on which the detector was originally evaluated, yielding seven detector-keyed groups. Crucially, the full pool of seven detectors is run on every input image: for an image drawn from detector~$d$'s evaluation benchmark, detector~$d$ - and any other detector covering the same manipulation family - operates within its intended regime, while the remaining detectors are applied out of scope, generating the cross-scope conflicts the framework is designed to resolve.

The \emph{out-of-distribution} collection comprises images drawn from datasets lying outside the training and evaluation data of \emph{every} pooled detector. Face manipulations are taken from Celeb-DF-v2~\cite{li2020celebdf}; whole-image generations from Community Forensics~\cite{park2025commforensics}, which spans a large and diverse set of recent generators; diffusion inpaintings from Adobe Firefly edits; GAN inpaintings from the six GAN inpainters of the IID-Net dataset~\cite{wu2022iidnet}; and splicing and copy-move forgeries from CASIA~v2.0~\cite{dong2013casia}. Authentic images are drawn from the corresponding source corpora - Celeb-DF-v2 source videos, Community Forensics reals, the originals underlying the Firefly edits, and CASIA~v2.0 authentic images - so that, within each manipulation family, genuine and manipulated content share the same data domain. Consequently, none of the detectors can rely on prior exposure to the test distribution, and the framework is assessed on its ability to arbitrate among detectors that are themselves operating out of distribution, closely reflecting the conditions encountered in real-world deployment.

\subsection{Experimental Setup and Metrics}
\label{sec:eval-setup}
Each agentic configuration (\cref{sec:eval-variants}) is driven by an MLLM backbone and is evaluated on both the in-distribution and out-of-distribution collections of \cref{sec:eval-data}. The decision threshold of $0.5$ applied to each detector’s normalised score is held fixed across all configurations - it is the natural midpoint of the $[0,1]$ score range and keeps the detector interface constant so that accuracy differences isolate the agentic components under study rather than reflect threshold calibration.

We report two metrics: binary accuracy is the proportion of images correctly judged manipulated or authentic; tool-arbitration accuracy examines, for the images the framework judges manipulated, whether the detector identified by the judge as the decisive detector is appropriate for the analyzed image.
The second metric is necessary because a high binary accuracy carries little weight if the framework reaches it for the wrong reason - attributing a manipulation to a detector whose competence does not cover the actual manipulation - and we therefore regard a verdict as fully correct only when both the binary decision and the arbitration are correct.

Quantifying arbitration correctness requires a ground-truth notion of which detector is appropriate for a given image.
Every test image is drawn from a source containing a single, known manipulation, so its true manipulation family is fixed by construction, without per-image annotation; each detector, in turn, declares the family or families that constitute its scope (\cref{tab:pool}).
We count the arbitrated detector as correct when its declared scope contains the image's true manipulation family (an \emph{in-scope} decision).
A manipulated image erroneously judged authentic yields no arbitration and counts against the metric.

\subsection{Framework Variants}
\label{sec:eval-variants}

\begin{table}[t]
  \centering
  \caption{The six variants of the agentic frameworks tested in our experiments. The first four form an ablation ladder on the Gemma-4-26B backbone; the last two vary the model. \cmark/\xmark{} indicate whether each component is present.}
  \label{tab:variants}
  \footnotesize
  \begin{tabular}{@{}lcccl@{}}
    \toprule
    Variant & \shortstack{MLLM\\agent} & \shortstack{Per-detector\\auditors} & \shortstack{Judge\\guidance} & Backbone \\
    \midrule
    Algorithmic fusion & \xmark & \xmark & \xmark & none \\
    Single-agent & \cmark & \xmark & \xmark & Gemma-4-26B \\
    Unguided judge & \cmark & \cmark & \xmark & Gemma-4-26B \\
    Complete framework & \cmark & \cmark & \cmark & Gemma-4-26B \\
    Complete framework & \cmark & \cmark & \cmark & GPT-4o \\
    Complete framework & \cmark & \cmark & \cmark & GPT-5 (judge only) \\
    \bottomrule
  \end{tabular}
\end{table}

We compare six configurations (\cref{tab:variants}), each run on both data collections, for a total of twelve runs.

The first four experiments constitute a progressive ablation study on the Gemma-4-26B backbone\footnote{Google DeepMind Gemma~4 26B A4B: a sparse mixture-of-experts model with 25.2B total parameters and $\sim$3.8B active per forward pass, released April 2026. \url{https://huggingface.co/google/gemma-4-26B-A4B-it}}, allowing us to quantify the individual contributions of triage, prompting, and evidence arbitration.

Algorithmic fusion is the no-agent floor: it groups detectors by declared manipulation scope, averages their normalised scores within each scope, and declares the image manipulated in the scope with the highest average among those with at least one detector above threshold. A single agent reinstates model reasoning but audits and arbitrates in one call, collapsing triage and arbitration; the unguided judge restores per-detector triage but omits the prompting axis; and the complete framework adds it.
The remaining two re-instantiate the complete framework with a different backbone - GPT-4o throughout, and GPT-5 as the judge with the auditors left on Gemma-4-26B.
All six share the same detector pool and score normalisation.

\subsection{Results}
\label{sec:eval-results}

\begin{figure}[t]
  \centering
  \includegraphics[width=\textwidth]{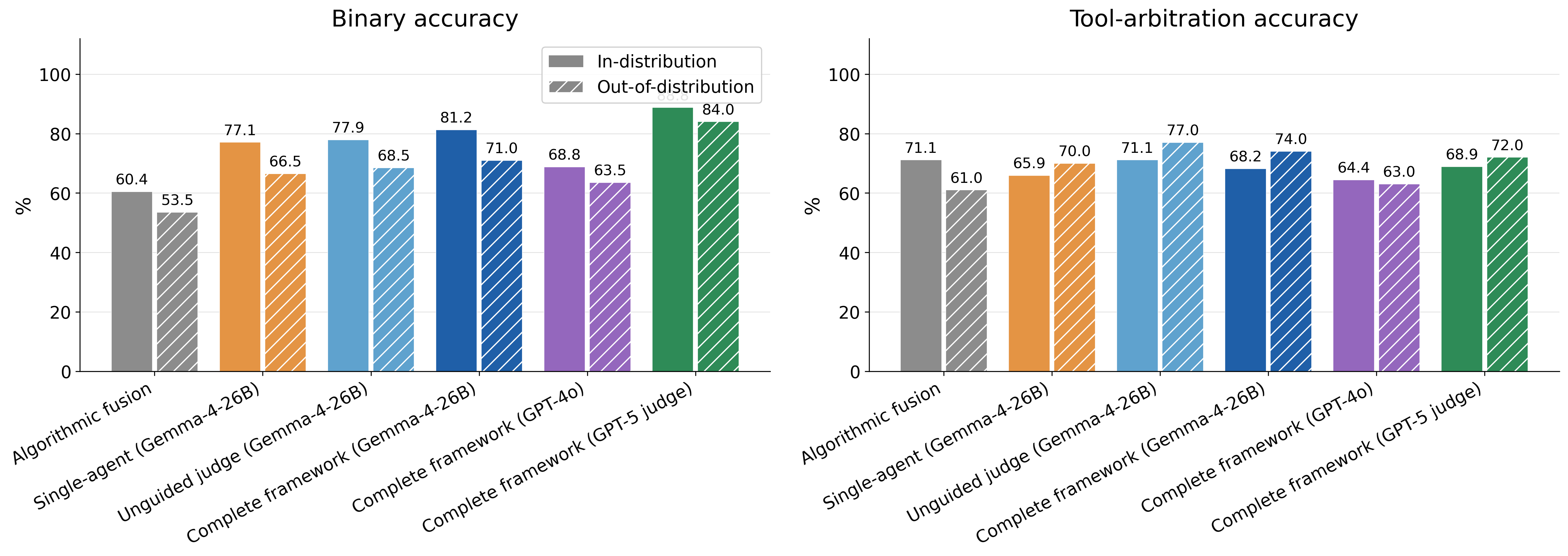}
  \caption{Binary accuracy (left) and tool-arbitration accuracy (right) for the six configurations (\cref{tab:variants}), in-distribution (solid) and out-of-distribution (hatched). Binary accuracy is the overall rate of correct authentic/manipulated verdicts; tool-arbitration accuracy is the in-scope detector-selection rate on images judged manipulated (\cref{sec:eval-setup}). }
  \label{fig:overview-acc}
\end{figure}

\begin{figure}[t]
  \centering
  \includegraphics[width=\textwidth]{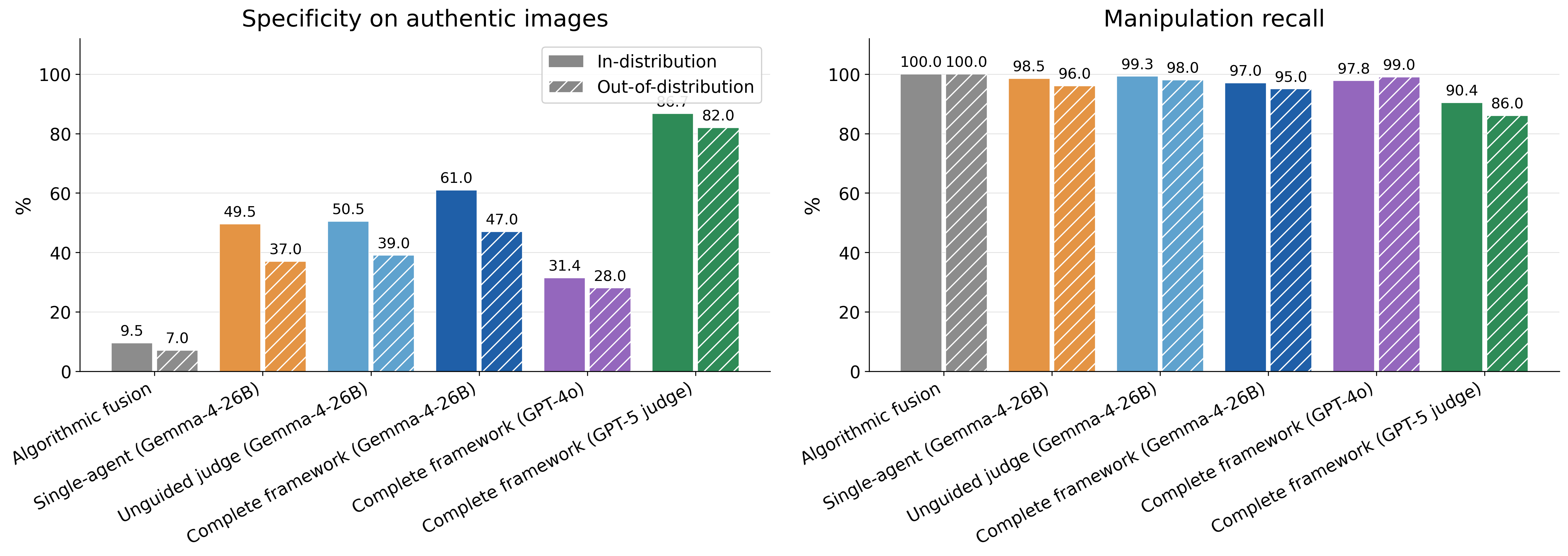}
  \caption{Specificity on authentic images (left) and recall on manipulated images (right), in-distribution (solid) and out-of-distribution (hatched). These are the two class-wise components of the binary accuracy of \cref{fig:overview-acc}: on the balanced sets their average is the binary accuracy. Manipulation recall is near-saturated everywhere, so the variation in binary accuracy---and the room for improvement---lies almost entirely in specificity on authentic images.}
  \label{fig:overview-breakdown}
\end{figure}

\Cref{tab:results} summarises the results of all twelve experimental configurations, while \cref{fig:overview-acc,fig:overview-breakdown} provide a visual overview. Since the relative trends are largely consistent across the two evaluation protocols, we focus the discussion on the out-of-distribution results, referring to the in-distribution setting only where differences are noteworthy. The discussion is organised around the four questions posed at the outset: the contributions of triage, prompting, and reasoning quality, together with the behaviour under distribution shift. For each, we consider two complementary perspectives: specificity on authentic images, and performance on manipulated images, assessed in terms of both manipulation recall and tool-arbitration accuracy.

\begin{table}[t]
  \centering
  \caption{Results (\%), in and out of distribution. Manip.\ is recall over manipulated images, Pristine is specificity on authentic images, Overall is binary accuracy, and Arb.\ is in-scope tool-arbitration accuracy. Best Overall per regime in bold.}
  \label{tab:results}
  \resizebox{\textwidth}{!}{%
  \setlength{\tabcolsep}{4pt}
  \begin{tabular}{@{}lcccccccc@{}}
    \toprule
    & \multicolumn{4}{c}{In-distribution} & \multicolumn{4}{c}{Out-of-distribution} \\
    \cmidrule(lr){2-5}\cmidrule(lr){6-9}
    Method & Manip. & Pristine & Overall & Arb. & Manip. & Pristine & Overall & Arb. \\
    \midrule
    Algorithmic fusion & 100.0 & 9.5 & 60.4 & 71.1 & 100.0 & 7.0 & 53.5 & 61.0 \\
    Single-agent (Gemma-4-26B) & 98.5 & 49.5 & 77.1 & 65.9 & 96.0 & 37.0 & 66.5 & 70.0 \\
    Unguided judge (Gemma-4-26B) & 99.3 & 50.5 & 77.9 & 71.1 & 98.0 & 39.0 & 68.5 & 77.0 \\
    Complete framework (GPT-4o) & 97.8 & 31.4 & 68.8 & 64.4 & 99.0 & 28.0 & 63.5 & 63.0 \\
    \textbf{Complete framework (Gemma-4-26B)} & 97.0 & 61.0 & 81.2 & 68.2 & 95.0 & 47.0 & 71.0 & 74.0 \\
    Complete framework (GPT-5 judge) & 90.4 & 86.7 & \textbf{88.8} & 68.9 & 86.0 & 82.0 & \textbf{84.0} & 72.0 \\
    \bottomrule
  \end{tabular}%
  }
\end{table}

\subsubsection{The Role of Triage.}
We isolate the contribution of per-detector auditors by comparing the single-agent configuration, which performs auditing and arbitration in a single pass, with the complete framework instantiated on the same Gemma backbone. The impact of triage is clear and unidirectional. Specificity on authentic images increases substantially, by approximately $10\%$, while manipulation recall remains effectively unchanged and tool-arbitration accuracy improves by a further 3--5\%. Triage thus yields a net benefit: it suppresses false positives on authentic content and sharpens detector attribution, with no meaningful loss in manipulation detection.

\subsubsection{The Role of Prompting.}
We isolate the effect of the judge's guidance by comparing the unguided judge with the complete framework, again using Gemma. Here the improvement comes with a mild trade-off: specificity on authentic images increases by roughly $8\%$, whereas manipulation recall and tool-arbitration accuracy each decrease slightly, by about 2--4\%. The guidance makes the judge mildly sceptical of the detectors' positive detector outputs, recovering additional authentic images at a modest cost on manipulated ones; the net effect on overall accuracy is nevertheless clearly favourable.

\subsubsection{The Role of Reasoning Quality.}

Reasoning quality is the dominant factor, as evidenced by the following three comparisons.

\emph{Reasoning versus none.} Relative to algorithmic fusion, the complete framework with Gemma delivers a substantial improvement on out of distribution experiments: specificity on authentic images increases from $7\%$ to $47\%$, and tool-arbitration accuracy from $61\%$ to $74\%$, at the expense of only a few percentage points of recall. Even a moderately capable reasoning agent is therefore what makes the detector pool practically usable.

\emph{Backbone capability.} The capability of the reasoning backbone is critical. The open-source Gemma~4 is more recent and empirically stronger than GPT-4o. The Gemma-based configuration outperforms the GPT-4o one on every metric: specificity ($47\%$ vs.\ $28\%$), overall accuracy ($71\%$ vs.\ $63.5\%$), and arbitration ($74\%$ vs.\ $63\%$), with manipulation recall already saturated in both cases. A more capable backbone therefore translates directly into a more accurate framework.

Upgrading only the judge. Motivated by this, we upgrade only the judge - the most demanding component of the pipeline - to GPT-5, while retaining Gemma as the auditor backbone. Even this partial change is decisive: specificity on authentic images increases from $47\%$ to $82\%$, manipulation recall decreases only slightly, and arbitration accuracy remains essentially unchanged. Overall accuracy rises to $84\%$, the highest among all configurations.

\subsubsection{Robustness to Distribution Shift.}
Distribution shift does not affect all aspects of the system uniformly. The degradation is largely confined to binary accuracy and, within it, to its two class-wise components: specificity decreases appreciably, whereas recall, already near saturation, changes only marginally. The main difficulty out of distribution thus lies in recognising unfamiliar authentic content rather than in detecting manipulations. By contrast, tool-arbitration accuracy does not deteriorate under distribution shift; for all agentic configurations it remains stable or improves slightly. The GPT-5 judge is the most robust overall, with accuracy decreasing only from $88.8\%$ to $84.0\%$.

Algorithmic fusion is the only clear outlier. Its tool-arbitration accuracy drops from $71.1\%$ for in-distribution to $61.0\%$ for out-of-distribution images, despite achieving the highest in-distribution accuracy among all configurations. This behaviour stems directly from the selection rule, which assigns each manipulation to the family with the highest average normalised score. In the in-distribution setting, every image is drawn from the benchmark on which its corresponding detector was originally evaluated, allowing that detector to consistently achieve the highest score and be correctly selected. Under distribution shift, however, this assumption no longer holds, leading to substantially more frequent misattributions. In contrast, the agentic configurations do not rely on such a regularity, which explains why their tool-arbitration accuracy remains remarkably stable across the two evaluation settings.

\subsection{Takeaways}
\label{sec:eval-takeaway}

Four observations emerge from our results.

First, every component of the agentic layer contributes positively to performance. Both per-detector auditing and conflict-aware arbitration yield consistent, albeit modest, gains, indicating that filtering unreliable evidence and reasoning over detector disagreements are beneficial in their own right.

Second, the quality of the reasoning model exerts a larger impact than any individual architectural component. Stronger judges do not replace triage or prompting redundant; rather, they exploit the additional structure and contextual information these mechanisms provide more effectively. Their advantage is most evident on authentic images, where they recover many cases that weaker backbones incorrectly classify as manipulated.

Third, the improvements introduced by the agentic layer arise primarily from a better handling of authentic content. Across configurations, gains in overall accuracy are consistently associated with substantial reductions in false positives, while changes in manipulation recall and tool-arbitration accuracy remain comparatively minor.

Fourth, manipulation recall is uniformly high, exceeding $85\%$ in all configurations and approaching saturation in most. Consequently, almost all variability in overall performance originates from the authentic class. This suggests that the central challenge of open-world image forensics is not the detection of manipulations per se, but the calibration of trust in forensic evidence and the reliable recognition of authentic content.

\section{Conclusion}
\label{sec:conclusion}
We have investigated open-world image forensics through a deliberately simple, training-free agentic framework that orchestrates a pool of pre-trained specialist detectors via per-detector triage and conflict-aware arbitration, and employed this framework as a testbed to identify the primary drivers of performance.
Our dissection yields a single overarching conclusion: in the open world, the main difficulty does not lie in identifying that an image has been manipulated - a task for which the detectors already achieve near-saturated recall - but in determining when a detector's output should be trusted. Nearly all variation in accuracy arises on the authentic class, where detectors operating outside their domain of competence generate spurious alarms.
The agentic layer attains its accuracy precisely by improving this trust decision: triage filters out detectors operating out of scope, and conflict-aware prompting evaluates the remaining evidence in relation to the image and each detector's declared domain. The quality of this decision, however, depends more on the reasoning capability of the model than on any particular architectural refinement: a stronger, better-calibrated judge improves specificity on authentic content far more than any architectural change, and upgrading the judge alone already suffices to obtain most of the benefit.
We therefore anticipate that further progress in agentic forensics will stem less from increasingly elaborate pipelines than from stronger reasoning capabilities and more accurately calibrated trust in individual detectors. At the same time, the framework's training-free and extensible design allows it to track the evolution of manipulation techniques by incorporating new detectors without additional training cost.

Relaxing the single-manipulation assumption to encompass composite manipulations, and replacing purely per-case reasoning with trust estimates calibrated from accumulated experience, constitute natural directions for future work.

\bibliographystyle{splncs04}
\bibliography{main}

\clearpage
\appendix
\raggedbottom

\renewcommand{\theHsection}{\Alph{section}}
\renewcommand{\theHsubsection}{\Alph{section}.\arabic{subsection}}

\crefname{section}{Appendix}{Appendices}
\Crefname{section}{Appendix}{Appendices}

\pdfbookmark[0]{Supplementary Material}{supplementary}
\begin{center}
  {\Large\bfseries Supplementary Material\par}
\end{center}
\vspace{0.5em}

\section{Overview}
\label{sec:supp-overview}

This supplement documents our full pipeline: the complete framework
of the main paper, with all components (per-detector auditing and judge guidance) and
a GPT-5 judge.
The pipeline's three stages---detector dispatch (Stage~0), per-detector auditing (Stage~1),
and conflict-aware judging (Stage~2)---are defined in the main paper; here we detail their
input-output mechanisms, showing some examples.
\Cref{sec:supp-walkthrough} follows the pipeline stage by stage on a single running example
(a pristine COCO photograph of zebras, hereafter the \emph{zebra case}), presenting each
stage's exact inputs and real outputs in order, from the dispatch table through the auditor
prompts and verdicts to the judge prompt and final output.

Prompts and outputs are reproduced verbatim from the deployed system.
Colour-coded annotations mark slot types throughout:
\textcolor{slotcol}{\bfseries red} marks content that is the same across all images but
varies across detectors (the \emph{per-detector} slot);
\textcolor{casecol}{\bfseries blue} marks content that changes for every image
(the \emph{per-case} slot).

\Cref{sec:supp-cases} presents two representative out-of-distribution (OOD) cases: one authentic image
and one manipulated image.
\Cref{sec:supp-variants} explains how the two ablated variants differ in their inputs.
All traces use Gemma-4-26B auditors and a GPT-5 judge.

\section{The Zebra Example: Stage-by-Stage Walkthrough}
\label{sec:supp-walkthrough}

\subsection{Stage~0: Detector Dispatch}
\label{sec:supp-dispatch}

The seven detectors described in the main paper run concurrently on the submitted image, without MLLM mediation/auditing. The verbatim traces below identify each
detector by its lowercase system identifier (e.g. \texttt{clipsid} = the CLIP-based detector from
Cozzolino \etal; \texttt{d3} = D\textsuperscript{3}; the other five are named as in the main paper).
\Cref{tab:supp-zebra-scores} shows the dispatch output for the zebra case.

\begin{table}[H]
  \centering
  \caption{Stage~0 dispatch output for the zebra case (authentic).
    \textbf{Score} = \texttt{is\_fake\_score} $\in [0,1]$: the unified manipulation score
    computed deterministically from each detector's native metric
    (0 = strong evidence authentic; 1 = strong evidence manipulated).
    \textbf{Fired} (\cmark) = \texttt{fired} = True iff Score~$>$~0.5, i.e., the
    detector reports manipulation; False (blank) = the detector stays silent.
    Both values are authoritative inputs to all downstream agents, and their variable
    names appear in all agent prompts and output traces below.}
  \label{tab:supp-zebra-scores}
  \scriptsize
  \begin{tabular}{@{}lllcc@{}}
    \toprule
    Detector & Scope & Native metric (raw value) & Score & Fired \\
    \midrule
    Cozzolino \etal  & Whole-image generation         & llr\_score = $-1.97$      & 0.122 & \\
    D\textsuperscript{3}   & Whole-image generation         & fake\_score = $0.29$      & 0.291 & \\
    Effort           & Face manipulation              & fake\_prob = $0.70$       & 0.699 & \cmark \\
    ForensicsAdapter & Face manipulation              & fake\_prob = $0.59$       & 0.585 & \cmark \\
    RADAR            & Diffusion inpainting           & tamper\_score = $1.00$    & 1.000 & \cmark \\
    NFA-ViT          & Diffusion / GAN inpainting     & fake\_score = $0.40$      & 0.399 & \\
    TruFor           & Inpainting, splicing/copy-move & integrity\_score = $0.53$ & 0.531 & \cmark \\
    \bottomrule
  \end{tabular}
\end{table}

\subsection{Stage~1: Per-Detector Auditor Agent}
\label{sec:supp-auditor}

\paragraph{Images.}
Each auditor call receives exactly the images its detector needs:
\begin{itemize}
  \item \textbf{Non-heatmap detectors} (Cozzolino \etal, D\textsuperscript{3}, Effort, ForensicsAdapter):
    only \textbf{Image~1} (the original photo).
  \item \textbf{Heatmap detectors} (RADAR, NFA-ViT, TruFor): \textbf{Image~1} plus
    \textbf{Image~2}, a tool-specific two-panel PNG generated by that detector's server:
    left panel = original overlaid with the detector's heatmap; right panel = pure jet
    heatmap.  Each heatmap auditor sees only its own tool's two-panel PNG, not the others'.
\end{itemize}

\begin{figure}[H]
  \centering
  \begin{minipage}[c]{0.35\textwidth}
    \centering
    \includegraphics[width=\linewidth]{figures/supp_zebra_original.jpg}\\[4pt]
    \small{(a) Image~1 — original photo, received by all seven auditors}
  \end{minipage}%
  \hfill
  \begin{minipage}[c]{0.62\textwidth}
    \centering
    \includegraphics[width=\linewidth]{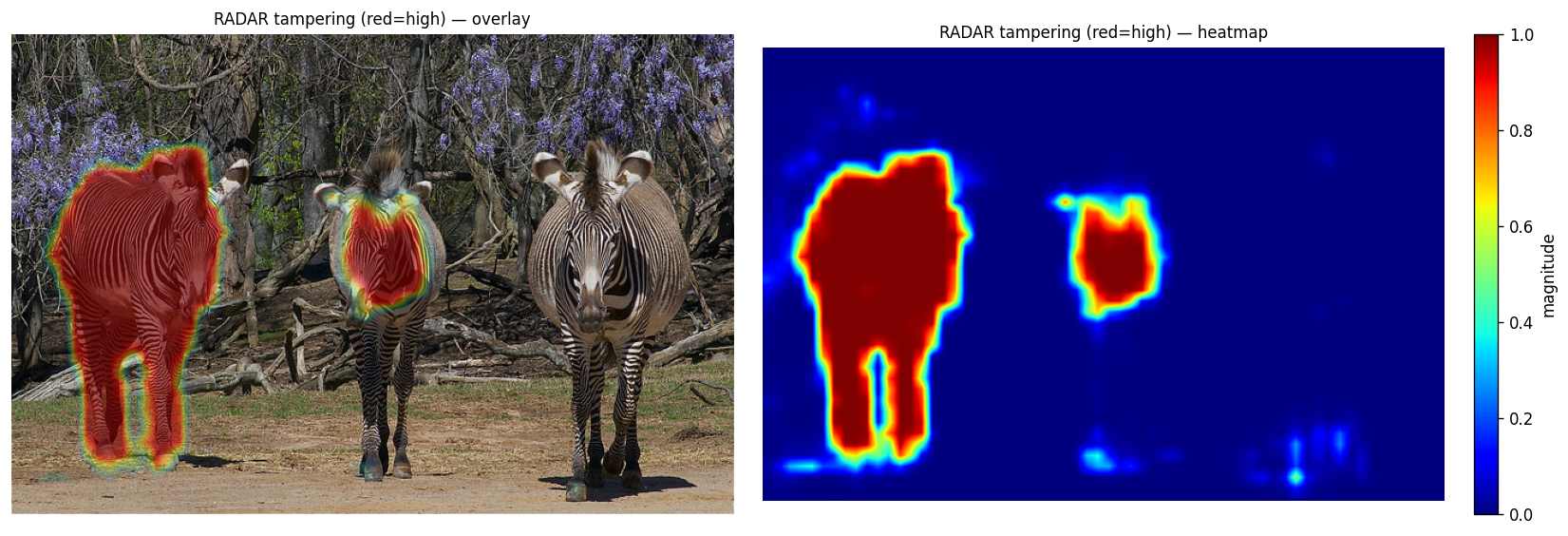}\\[4pt]
    \small{(b) Image~2 for the RADAR auditor — two-panel PNG: left = original
      with RADAR's overlay; right = pure jet heatmap}
  \end{minipage}
  \caption{The two images sent to the RADAR auditor for the zebra case.
    Non-heatmap auditors (Cozzolino \etal, D\textsuperscript{3}, Effort, ForensicsAdapter) receive only
    Image~1.  Each heatmap auditor (RADAR, NFA-ViT, TruFor) additionally receives its own
    two-panel PNG; these are separate per-tool files, not the composite panel the judge sees.}
  \label{fig:supp-zebra-auditor-inputs}
\end{figure}

\Cref{fig:supp-zebra-auditor-inputs} shows Image~1 and RADAR's Image~2 — the two images
that the RADAR auditor receives.

\paragraph{Prompt.}
\Cref{lst:auditor} is the verbatim prompt sent to the \textbf{RADAR} auditor.
\textcolor{slotcol}{\bfseries Red}: the \textsc{Tool Card} block — RADAR's identity, scope, threshold, and heatmap description.
This block is different for each of the seven detectors but stays fixed across all images.
\textcolor{casecol}{\bfseries Blue}: the \textsc{This Run} block — the dispatch output specific to this image.
This block changes for every image (and also reflects the detector's native output format).
The TASK, YOUR REASONING, and OUTPUT sections are identical for all seven auditors on all images.
Non-heatmap detectors (Cozzolino \etal, D\textsuperscript{3}, Effort, ForensicsAdapter) receive the same
prompt structure but without the heatmap instructions in STEP~2.

\FloatBarrier
\begin{lstlisting}[caption={Verbatim prompt sent to the RADAR auditor for the zebra case.
  \textcolor{slotcol}{\bfseries Red}: per-detector tool card.
  \textcolor{casecol}{\bfseries Blue}: per-case dispatch output.
  The remainder is the shared fixed template.},
  label={lst:auditor},captionpos=t]
Respond with ONLY a valid JSON object. No prose, no markdown, no code fences.

You are the per-tool auditor agent for ONE specific detector in a multi-agent forensic pipeline.

-------------- TASK --------------
Decide whether THIS detector's output is credible evidence on THIS image, and produce one ``audit_verdict``.  Your verdict is the gatekeeper -- ``invalid`` removes the tool from every downstream stage (it will not be considered by any later agent); a live verdict (e.g. ``valid``) propagates the tool's output downstream.

The verdict integrates three checks IN ORDER, with check 1 as a HARD PREREQUISITE for checks 2 and 3:

  1. SCOPE   -- does the image actually contain the kind of content this tool is designed to detect?  This is the prerequisite: **when SCOPE fails, the verdict is ``invalid`` regardless of anything checked later** (a tool applied outside its scope produces a meaningless score, no matter how the heatmap looks, no matter how far the score is from threshold, no matter what any additional rule below says).
  2. SIGNAL  -- if SCOPE passed, is the is_fake_score interpretable (not stuck right at threshold), and (for heatmap tools) does the spatial pattern coherently support the score?
  3. CONTEXT -- if an additional verdict-rule block is rendered below, use it to refine the choice AMONG THE LIVE VERDICTS -- it does NOT override check 1.  ("Live" means the verdicts that propagate the tool downstream; a refined choice within them just tells downstream how much weight to put on the tool.)

@-------------- TOOL CARD --------------
Tool: RADAR

Designed to detect:
  - Localised diffusion-based image manipulations: object insertion, object removal via background inpainting, and region-level edits produced by text-conditioned latent-diffusion inpainters. Produces both an image-level score and a per-pixel tampering probability map.

Decision metric threshold:
  - tamper_score > 0.5 -> predicted tampered; < 0.5 -> predicted clean. The score is the mean of the top 1 % highest pixel values in the tampering probability map (paper-defined image-level aggregation).
How to read the output: tamper_score is the paper-defined image-level aggregation over the 1 % most-suspicious pixels in the tampering map. The continuous tampering map explains where the evidence is spatially located; cross-validate by checking whether bright regions correspond to a semantically plausible edit target in the original image.
Heatmap semantics: Two-panel figure.  LEFT = original image.  RIGHT = jet-coloured tampering probability map (red = high tampering probability, blue = clean).  Use the RIGHT panel to localise the inpainted region; verify against the original image.
Attached image (Image 2) is a 2-panel PNG for this tool: left = original
with overlay heatmap; right = pure heatmap. Hot (red/orange) regions indicate the tool's suspected manipulation locations.@
!-------------- THIS RUN --------------
Tool ID: radar
Raw tool output (sanitised; file paths removed): {"is_tampered": true,
  "runtime_s": 0.88, "schema": "1.2", "status": "ok",
  "tamper_score": 1.0, "tool": "radar"}

Normalized decision signal (computed by code, unified across all tools):
  is_fake_score = 1.0000   (range [0, 1]; threshold 0.5)
                  -> 0 means strong evidence REAL; 1 means strong evidence FAKE
  fired         = True   (= is_fake_score > 0.5)
  raw source    = tamper_score=1.0000  (the tool's native score before normalization)

These two values (``is_fake_score`` and ``fired``) are AUTHORITATIVE and will be re-emitted by the code unchanged.  Your job is NOT to recompute them -- your job is to judge whether the verdict they imply is CREDIBLE (per the TASK above).!

-------------- YOUR REASONING (follow in order) --------------

STEP 1 -- Scope-input applicability  (the TASK's hard prerequisite)
   Look at Image 1 (the original photo).  Does it contain the kind of content this tool is designed to detect (see "Designed to detect" above)?  When the tool's scope requires content that is absent from this image (e.g. a face-deepfake detector applied to a landscape or to animals rather than human faces; a local-edit detector applied to clearly synthetic whole-image content), the tool is not applicable here -- its score is meaningless on this image.  Per the TASK, this case yields ``audit_verdict = "invalid"`` and any later step (SIGNAL, CONTEXT) cannot rescue it.  Otherwise proceed to STEP 2.

STEP 2 -- Output interpretation
   Given is_fake_score = 1.0000 and threshold 0.5:
   - How far is it from the threshold?  ~0.5 = weak signal; near 0 or 1 = strong.
   - "fired = True" means the tool claims the image IS manipulated.
   - Look at the attached heatmap (Image 2).  Describe the spatial pattern of the hot (red/orange) region neutrally:
       * Concentrated on a specific object/region in Image 1 -> potential support for the fire.
       * Scattered isolated pixels with no clear spatial structure -> potential noise pattern; raise the suspicion level.
       * Uniformly high across the whole heatmap -> low-information pattern (saturated everywhere ~= "no localised anomaly"); raise the suspicion level.
   - Ask whether the hot region corresponds to ANY plausible manipulation target visible in Image 1.  A clearly-named target (a specific object, a clean region boundary) is potential support for the fire; absence of a nameable target raises the suspicion level.
   - Do NOT pre-judge specific visual patterns as "always FP" (e.g. activation on object edges, image borders, or fabric texture).  Such patterns CAN host real manipulations.  If a data-driven FP signal for THIS tool on a property visible in Image 1 is supplied elsewhere in this prompt, defer to that signal rather than inventing hard-coded rules of your own here.

STEP 3 -- Credibility verdict
   You are this tool's gatekeeper.  Your verdict here determines whether the tool's output is forwarded to downstream stages or dropped from the case entirely.  Integrate your STEP 1 and STEP 2 observations and pick the verdict that best describes this tool's output on this image.

   What each value MEANS to downstream stages:

     "valid"   -- the tool's output is credible evidence (in either direction).  A credible NEGATIVE (low is_fake_score, fired=false) is just as useful as a credible positive; downstream agents read both faithfully.

     "invalid" -- the tool's output is not credible enough to count as evidence (e.g. out-of-scope content, uninterpretable signal near the threshold, or STEP 2 suspicion signals that compoundingly undermine the score).  Downstream agents will exclude it.

   When an additional verdict-rule block is rendered above (e.g. an extended enum or tier-specific guidance), apply the more specific rule.

-------------- OUTPUT (strict JSON, no prose, no markdown) --------------
{
  "tool_id": "radar",
  "audit_verdict": <one of the values defined by the verdict rules above>,
  "reasoning": "<concise but specific; cite which step led to the verdict; <=150 words>"
}
\end{lstlisting}

\paragraph{Outputs.}
\Cref{lst:zebra-audit} gives the verbatim outputs from all seven auditor calls.
Effort and ForensicsAdapter report a manipulation\footnote{%
  In the verbatim traces, \texttt{fired = True} (score $> 0.5$) denotes a detector
  reporting a manipulation; \texttt{audit\_verdict = "valid"} / \texttt{"invalid"}
  corresponds to \emph{retained} / \emph{excluded} in the main paper's language.}
at 0.70 and 0.59 respectively, yet are immediately
invalidated because STEP~1 finds no single dominant human face in the image.
The remaining five detectors --- including RADAR at score~1.0 --- are retained and
forwarded to the judge.

\begin{lstlisting}[caption={Stage~1 auditor outputs for the zebra case, verbatim from the
  trace (\texttt{is\_fake}: code-computed normalised score; \texttt{fired}: \texttt{is\_fake}$>$0.5;
  the native metric follows in brackets). Each verdict line is followed by that auditor's
  reasoning. Effort and ForensicsAdapter are invalidated at STEP~1 for scope (no human
  face); the other five are retained.},captionpos=t,label={lst:zebra-audit}]
clipsid -> valid   [is_fake=0.122 fired=False llr_score=-1.9724]
  STEP 1: The tool is designed for whole-image AI detection, which is applicable to this image of zebras. STEP 2: The signal is clear and interpretable; the is_fake_score of 0.1221 is well below the 0.5 threshold, and the llr_score of -1.9724 indicates a strong prediction that the image is real. STEP 3: Since the tool is within scope and provides a decisive, non-ambiguous signal, the output is credible evidence for downstream agents.

d3 -> valid   [is_fake=0.291 fired=False fake_score=0.2913]
  STEP 1: The image contains a whole-scene composition of zebras, which falls within the scope of a whole-image AI-generation detector (D3). STEP 2: The is_fake_score of 0.2913 is well below the 0.5 threshold, indicating a clear signal that the image is predicted to be real. The score is not stuck at the threshold, providing a confident negative result. STEP 3: Since the scope is correct and the signal is interpretable and clear, the output is credible evidence for downstream agents to consider as a 'real' classification.

nfavit -> valid   [is_fake=0.399 fired=False fake_score=0.3985]
  STEP 1: The tool is designed to detect inpainting/local manipulations, which is applicable to the zebra image. STEP 2: The is_fake_score is 0.3985, which is below the 0.5 threshold, indicating a 'real' classification. The heatmap shows diffuse, low-magnitude activations (cyan/blue) across the zebra bodies rather than high-confidence red/orange regions. While the signal is not extremely strong, it is not stuck exactly at the threshold, and the lack of a concentrated, high-magnitude heatmap is consistent with the tool's decision that no significant inpainting was detected. Therefore, the output is a credible negative signal.

radar -> valid   [is_fake=1.000 fired=True tamper_score=1.0000]
  STEP 1: The image contains zebras, which are suitable subjects for detecting localized diffusion-based manipulations like object insertion or inpainting. STEP 2: The is_fake_score is 1.0, a very strong signal far from the 0.5 threshold. The heatmap (Image 2) shows highly concentrated, high-magnitude (red) regions specifically localized on the bodies of the first two zebras. This spatial coherence—where the 'tampering' is mapped to specific semantic objects rather than being scattered noise or uniform across the image—provides strong support for the tool's claim of localized manipulation. STEP 3: The signal is interpretable and spatially consistent with the tool's design, making it credible evidence.

trufor -> valid   [is_fake=0.531 fired=True integrity_score=0.5314]
  STEP 1: The image contains real animals (zebras) in a natural setting, which is within the scope of TruFor's ability to detect local manipulations like splicing or inpainting. STEP 2: The is_fake_score (0.5314) is very close to the 0.5 threshold, indicating a weak signal. However, the heatmap (Image 2) shows highly concentrated, high-magnitude (red) regions specifically localized on the bodies of the zebras. This spatial coherence—where the anomaly is not scattered noise but follows the semantic structure of the subjects—provides interpretable support for the score. While the signal is weak, the pattern is not uninterpretable or nonsensical, making the output credible evidence for downstream agents to evaluate.

effort -> invalid   [is_fake=0.699 fired=True fake_prob=0.6993]
  STEP 1: The tool is designed to detect face-manipulation deepfakes on a single human face presented as the dominant subject. The provided image contains zebras, not human faces. Therefore, the image is outside the tool's scope, making the output meaningless. Per the task instructions, a scope failure results in an 'invalid' verdict regardless of the score.

forensicsadapter -> invalid   [is_fake=0.585 fired=True fake_prob=0.5851]
  STEP 1: The tool is designed specifically for face forgery detection on a single human face (portrait/headshot). The provided image contains zebras, not human faces. Therefore, the image is outside the tool's scope, making the output meaningless. Per the task instructions, a scope failure results in an 'invalid' verdict regardless of the score.
\end{lstlisting}

\subsection{Stage~2: Final Judge Agent}
\label{sec:supp-judge}

\begin{figure}[H]
  \centering
  \includegraphics[width=\textwidth]{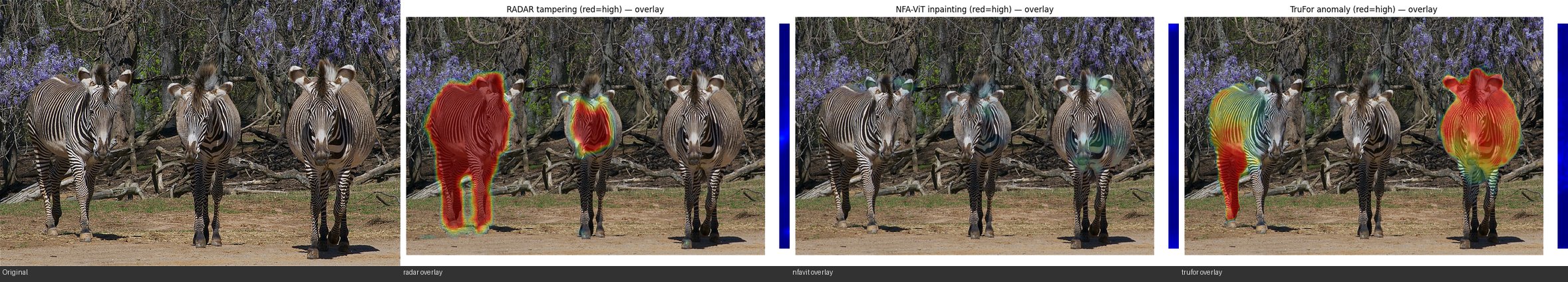}\\[4pt]
  \small{(a) Image~2 — composite overlay panel (sent to judge)}\\[8pt]
  \includegraphics[width=\textwidth]{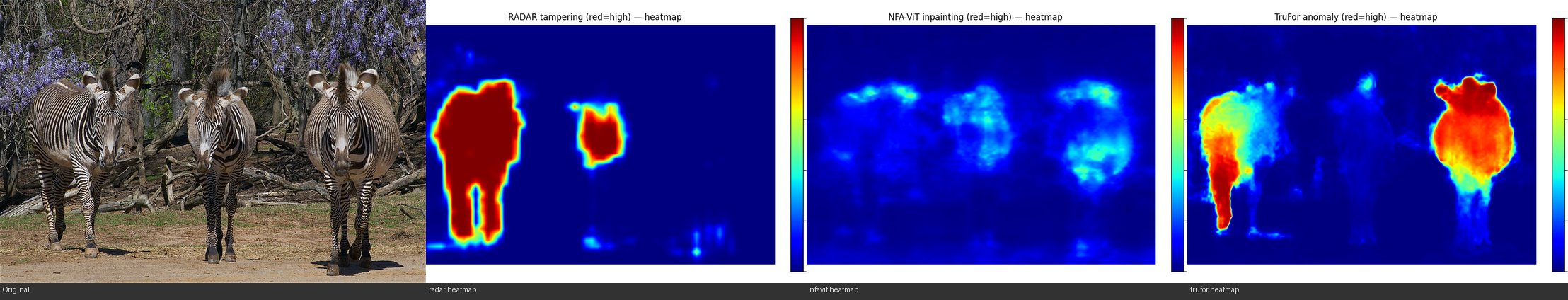}\\[4pt]
  \small{(b) Image~3 — composite raw heatmap panel (sent to judge)}
  \caption{Images~2 and~3 received by the judge for the zebra case.
    In each panel, the top-left tile is the original photo for reference;
    the remaining three tiles show outputs from RADAR, NFA-ViT, and TruFor.
    These composite panels are \emph{only} sent to the judge — individual auditors
    each see their own per-tool two-panel PNG (Fig.~\ref{fig:supp-zebra-auditor-inputs}b).}
  \label{fig:supp-zebra-judge-imgs}
\end{figure}

\paragraph{Images.}
The judge receives three images: Image~1 (original photo, same as the auditors' input),
Image~2 (composite overlay panel), and Image~3 (composite raw heatmap panel).
These composite panels are assembled from the three heatmap detectors' individual outputs
after all auditor calls complete.
\Cref{fig:supp-zebra-judge-imgs} shows Images~2 and~3 for the zebra case.

\paragraph{Prompt.}
The judge prompt has a large fixed body (TASK, REASONING PROTOCOL STEP~1--3,
DECISION instructions, and OUTPUT format) and two sections that change per case,
shown in \textcolor{casecol}{\bfseries blue} in \cref{lst:judge}.
The \textsc{Tool Cards} block (shown in \textcolor{slotcol}{\bfseries red}) lists one
card per active detector — it is per-detector, but the set of active detectors varies
per case as some are invalidated (here, two face detectors were excluded).
The \textsc{Audit Evidence} block (blue) carries each retained auditor's verbatim
verdict and reasoning — fully per case; these are the Stage~1 auditor outputs listed
in full in \cref{lst:zebra-audit}.

\FloatBarrier
\begin{lstlisting}[caption={Verbatim judge prompt for the zebra case.
  \textcolor{slotcol}{\bfseries Red}: TOOL CARDS block (per active detector).
  \textcolor{casecol}{\bfseries Blue}: AUDIT EVIDENCE block (per case).
  The remainder is the shared fixed template.},
  label={lst:judge},captionpos=t,basicstyle=\ttfamily\tiny]
You are the final judge agent.

-------------- TASK --------------
Integrate your own visual inspection of Image 1, the per-tool auditor agents' verdicts, the tool cards, and the composite heatmap panels into a SINGLE forensic verdict for this image.

A core part of the integration is identifying and resolving DISAGREEMENTS among detectors -- multiple detectors with overlapping or different scopes often produce inconsistent verdicts on the same image; your job is to read those inconsistencies and decide what they mean.

You do NOT re-do the per-tool auditor agents' work.  The auditor agents already decided which tools' outputs are credible on this image (you only see the surviving ones).  You integrate what you have.

Your output:
  - is_manipulated   (true or false)
  - manipulation_locality / manipulation_region
  - arbitrated_tools (the tools whose evidence drove the verdict)
  - decision_path    (short ordered bullets naming the steps you took)
  - reasoning        (written FIRST as a narrative, then the structured fields fall out from it)

You receive:
  - Image 1: the original photo under examination.
  - Image 2: composite OVERLAY panel -- for each heatmap-producing tool, its heatmap is blended on top of the original.  Tile-by-tile, labeled by tool name.  Top-left tile is the unaltered original for reference.
  - Image 3: composite RAW heatmap panel -- pure jet heatmaps from the same tools in the same tile layout, labeled.  Top-left tile is the original.
  - Tool cards: one card per ACTIVE tool -- its scope_in, threshold, and signal interpretation.  These define what each tool is designed to detect; you'll use them to reason about which detectors should agree vs disagree on this image.  "Active" here means the per-tool auditor agent did not invalidate the tool; invalidated tools do NOT appear.
  - Per-tool audit table: verdicts from the per-tool auditor agents (one auditor per detector; each examined a single tool's output in isolation).  ``audit_verdict in {valid, invalid}``; tools the auditor agents invalidated do NOT appear in this table.

-------------- REASONING PROTOCOL --------------

STEP 1 -- Independent semantic visual inspection of Image 1
   Describe what the image depicts and note anything visually inconsistent:
   - Lighting and shadow consistency across objects.
   - Texture uniformity (does grain/noise match across regions?).
   - Perspective and scale (do foreground/background elements relate correctly?).
   - Object boundary quality (sharp edges where they should not be, missing shadows, mismatched depth-of-field).
   - For potential AI generation: hand/finger anomalies, unnatural skin, implausible backgrounds, malformed text.
   Your semantic observation is INDEPENDENT EVIDENCE -- it can confirm or override tool verdicts.

STEP 2 -- Identify potential conflicts among detectors
   Every surviving tool has two attributes that together determine whether its output is consistent with another tool's output:

     - SCOPE  (from the tool card's scope_in): the kind of manipulation this tool is designed to detect.
     - OUTPUT (from the audit table): ``fired`` = claims manipulation; silent = claims no manipulation in its scope.

   A "conflict" exists when two or more tools' (scope, output) pairs cannot be JOINTLY true on this image.  The governing principle:

     At most ONE scope can be the true scope of a manipulation, if any manipulation is present at all.  Therefore, detectors whose scopes overlap but whose outputs disagree, OR detectors whose scopes do not overlap but whose outputs both fire, cannot all be right on this image.  Detectors that agree within a shared scope, or that all stay silent across scopes, are jointly consistent and need no resolution.

   Walk through the surviving tools, compare their (scope, output) pairs across one another, and list the conflict groups you find. The principle above is the criterion, not an exhaustive catalogue -- derive the conflict-vs-no-conflict distinction yourself for each combination you encounter.  Briefly state the conflicts you identified in your reasoning output (this is your audit trail).  Do NOT attempt resolution yet -- that is STEP 3.

STEP 2b -- Reason about per-tool reliability on THIS image
   Each surviving tool is a statistical classifier whose paper-reported accuracy is meaningfully below 100%.  A tool that fires is making a probabilistic claim -- sometimes a correct detection within its scope, sometimes a false alarm where the tool over-triggered on image content that resembled its training distribution.  A silent tool is correspondingly sometimes a missed detection.  When you weigh a surviving tool's verdict, you are weighing the joint plausibility of "true detection on this image" against "this is one of the tool's known failure modes".

   The lens for reasoning about whether THIS tool's output on THIS image is credible evidence:

     - Scope vs image content (your STEP 1 visual reading): does Image 1 plausibly contain the manipulation type the tool was designed to detect?  A face-manipulation detector firing on a landscape with no visible face has no place for a true positive to be; a whole-image AI-generation detector firing on a photograph with clear camera noise and consistent lighting is making a claim that the scene's natural evidence contradicts.
     - Score magnitude relative to the decision boundary (from the tool card's threshold): a score sitting close to the boundary ridge is qualitatively different from a saturated score -- borderline values are more often false alarms than confident detections.
     - Heatmap shape, for heatmap-producing tools (Images 2 and 3): a spatial pattern that lands on an attacker-plausible region (a single coherent object, a clear inpainting target, a face) is qualitatively different from a pattern that scatters across natural-texture content (water ripples, foliage, dense repeating patterns, sky) where spurious activations are well-known in the forensic literature.
     - Cross-tool corroboration within the same scope: a positive backed by another in-scope peer is more likely a true detection than a lone positive surrounded by silent peers sharing the same scope.

   These cues are NOT a checklist with a scored outcome -- they are the lens through which you re-read each surviving tool's output.  Your reading of each tool ("this output is credible evidence" vs "this output is likely a false alarm on this image") is what STEP 3 then integrates with your STEP 1 visual observations to resolve the STEP 2 conflicts.

STEP 3 -- Resolve conflicts and integrate into the final verdict
   For each conflict you listed in STEP 2, decide which direction survives.  Inputs available for your decision:

     - Tool cards' scope_in -- which detector's design scope actually matches the content visible in Image 1 (per your STEP 1 visual inspection)?
     - Composite heatmap panels (Image 2 overlay, Image 3 raw) -- do firing heatmap detectors highlight the SAME image structure (multi-tool spatial corroboration) or DIFFERENT structures (score-level agreement without spatial agreement)?
     - Per-tool auditor reasoning (in the audit table) -- what each auditor saw and why it gave its verdict.

   Integrate the surviving evidence + your STEP 1 visual observations into the final verdict.  When the surviving tools and your visual observations point in different directions, you decide how to weigh them given the inputs available to you.

@TOOL CARDS (per active tool):
  clipsid (ClipSID):
    Detects: Whole-image classification: is the entire image AI-generated content?.
    Threshold: llr_score: signed scalar -- negative -> predicted real, positive -> predicted AI-generated (decision boundary at 0).
    Signal: llr_score is a signed scalar.  Sign = predicted class; magnitude = relative distance from the decision boundary. It is NOT a calibrated probability.
  d3 (D3):
    Detects: Whole-image AI-generation detection (GAN and diffusion)..
    Threshold: fake_score > 0.5 -> predicted AI-generated; < 0.5 -> predicted real (fixed decision boundary).
    Signal: fake_score is a sigmoid scalar.  Above 0.5 -> the model detected AI-generation artifacts; below 0.5 -> feature distributions look natural.  Magnitude reflects the model's confidence, not a calibrated probability.
  radar (RADAR):
    Detects: Localised diffusion-based image manipulations: object insertion, object removal via background inpainting, and region-level edits produced by text-conditioned latent-diffusion inpainters.  Produces both an image-level score and a per-pixel tampering probability map..
    Threshold: tamper_score > 0.5 -> predicted tampered; < 0.5 -> predicted clean.  The score is the mean of the top 1 % highest pixel values in the tampering probability map (paper-defined image-level aggregation).
    Signal: tamper_score is the paper-defined image-level aggregation over the 1 % most-suspicious pixels in the tampering map.  The continuous tampering map explains where the evidence is spatially located; cross-validate by checking whether bright regions correspond to a semantically plausible edit target in the original image.
  nfavit (NFA-ViT):
    Detects: Image inpainting localization -- GAN-based, traditional, and diffusion-based inpainting alike.  Produces both an image-level score and a per-pixel segmentation mask of the inpainted region..
    Threshold: fake_score > 0.5 -> inpainting detected (paper-derived decision boundary).  Values near 0.5 are uncertain; > 0.8 indicates high-confidence inpainting.
    Signal: fake_score is the top-1 % mean of pred_mask, the pixel-level segmentation output of a trained decoder head (not a saliency / gradient explanation).  A high fake_score with no compact heatmap region is diagnostically inconsistent and should raise suspicion of a false positive.
  trufor (TruFor):
    Detects: Local-manipulation localization: splicing, copy-move, inpainting (including diffusion-based).  Produces both an image-level score and a pixel-level anomaly map..
    Threshold: integrity_score is in [0, 1]; decision boundary at 0.5 (higher -> more likely tampered).
    Signal: integrity_score -> higher means more likely tampered. anomaly_max = peak local suspicion; anomaly_mean = background floor.  A LOW integrity score does NOT automatically rule out manipulation if the anomaly map shows a concentrated hot region at a semantically plausible tamper location.@

-------------- DECISION --------------
After working through the reasoning protocol above, integrate to the final verdict.

WRITE ``reasoning'' FIRST as a step-by-step narrative.  Walk through every reasoning element rendered above IN THE ORDER IT APPEARED: each section / step / rule.  For each, name what you considered, name the rule you applied, and name the surviving tool(s).  Once the narrative is complete, the remaining fields fall out:

- arbitrated_tools = a faithful transcription of the conclusion you just wrote -- the specific tool(s) your reasoning named as the DECISIVE evidence for the verdict, listed most-decisive FIRST. This is NOT a re-listing of every tool that fired, nor every tool the per-tool auditor agent left live; it is exactly the driver(s) your final reasoning step settled on.  When your reasoning prioritized a scope or tool ("I favor the local detectors", "X is the most credible", "the face evidence is primary"), those favored tool(s) are the arbitrated ones and the first element is the one you treated as primary.  A tool you described only as corroborating / supporting / consistent-with / secondary stays OUT -- it shaped your thinking but did not drive the verdict.  Non-empty when is_manipulated=true; empty when is_manipulated=false.
- is_manipulated = true iff ``arbitrated_tools'' is non-empty AND the winning evidence is not contradicted by the observations available to you.
- manipulation_locality = "global" (whole-image AI-gen-style) or "local" (a specific region) or "n/a" (pristine).
- manipulation_region = brief textual description of where the edit is, if local; null otherwise.
- decision_path = ordered list of short bullet strings naming the reasoning steps you walked.  Each bullet <= 25 words.

!AUDIT EVIDENCE:
  clipsid: valid fired=False score=0.122 -> STEP 1: The tool is designed for whole-image AI detection, which is applicable to this image of zebras. STEP 2: The signal is clear and interpretable; the is_fake_score of 0.1221 is well below the 0.5 threshold, and the llr_score of -1.9724 indicates a strong prediction that the image is real. STEP 3: Since the tool is within scope and provides a decisive, non-ambiguous signal, the output is credible evidence for downstream agents.
  d3: valid fired=False score=0.291 -> STEP 1: The image contains a whole-scene composition of zebras, which falls within the scope of a whole-image AI-generation detector (D3). STEP 2: The is_fake_score of 0.2913 is well below the 0.5 threshold, indicating a clear signal that the image is predicted to be real. The score is not stuck at the threshold, providing a confident negative result. STEP 3: Since the scope is correct and the signal is interpretable and clear, the output is credible evidence for downstream agents to consider as a 'real' classification.
  radar: valid fired=True score=1.000 -> STEP 1: The image contains zebras, which are suitable subjects for detecting localized diffusion-based manipulations like object insertion or inpainting. STEP 2: The is_fake_score is 1.0, a very strong signal far from the 0.5 threshold. The heatmap (Image 2) shows highly concentrated, high-magnitude (red) regions specifically localized on the bodies of the first two zebras. This spatial coherence--where the 'tampering' is mapped to specific semantic objects rather than being scattered noise or uniform across the image--provides strong support for the tool's claim of localized manipulation. STEP 3: The signal is interpretable and spatially consistent with the tool's design, making it credible evidence.
  nfavit: valid fired=False score=0.399 -> STEP 1: The tool is designed to detect inpainting/local manipulations, which is applicable to the zebra image. STEP 2: The is_fake_score is 0.3985, which is below the 0.5 threshold, indicating a 'real' classification. The heatmap shows diffuse, low-magnitude activations (cyan/blue) across the zebra bodies rather than high-confidence red/orange regions. While the signal is not extremely strong, it is not stuck exactly at the threshold, and the lack of a concentrated, high-magnitude heatmap is consistent with the tool's decision that no significant inpainting was detected. Therefore, the output is a credible negative signal.
  trufor: valid fired=True score=0.531 -> STEP 1: The image contains real animals (zebras) in a natural setting, which is within the scope of TruFor's ability to detect local manipulations like splicing or inpainting. STEP 2: The is_fake_score (0.5314) is very close to the 0.5 threshold, indicating a weak signal. However, the heatmap (Image 2) shows highly concentrated, high-magnitude (red) regions specifically localized on the bodies of the zebras. This spatial coherence--where the anomaly is not scattered noise but follows the semantic structure of the subjects--provides interpretable support for the score. While the signal is weak, the pattern is not uninterpretable or nonsensical, making the output credible evidence for downstream agents to evaluate.

Excluded by the per-tool auditor agent (do not consider): effort, forensicsadapter!

-------------- OUTPUT (strict JSON, no prose, no markdown) --------------
{
  "reasoning":             "<<=200 words; step-by-step narrative in
                             the order the prompt rendered the reasoning elements; cite the rule and the surviving tool(s) for each>",
  "is_manipulated":        true or false,
  "manipulation_locality": "global" or "local" or "n/a",
  "manipulation_region":   "<brief region description or null>",
  "arbitrated_tools":      ["<tool_id>", ...],
  "decision_path":         ["<step1>", "<step2>", ...]
}
\end{lstlisting}

\paragraph{Output.}
\Cref{lst:judge-out} shows the real GPT-5 output for the zebra case.\footnote{%
  \texttt{arbitrated\_tools} = the decisive detector of the main paper (most-decisive first).}

\begin{lstlisting}[caption={Real GPT-5 judge output for the zebra case (verbatim from trace).},
  label={lst:judge-out},captionpos=t]
reasoning: STEP 1 — Visual inspection: Photo of three zebras walking in daylight.  Lighting and shadows are consistent under each zebra; textures/noise uniform; perspective and boundaries natural; no seams or depth-of-field mismatches.  No obvious local edits.
  STEP 2 — Conflicts: Whole-image detectors (ClipSID, D3) both predict real (silent).  Local tools disagree: RADAR fires strongly on the left zebra and center chest; TruFor barely above threshold and highlights mostly the right zebra; NFA-ViT is below threshold with diffuse maps.  These constitute intra-scope conflicts (only one local-edit scope can be true).
  STEP 2b — Reliability on this image: RADAR's saturated score but object-sized blobs align with high-contrast zebra stripes — a known false-positive surface.  TruFor's score is near the decision boundary and its hotspot is in a different place, weakening credibility. NFA-ViT's negative with no compact mask argues against inpainting. Lack of spatial corroboration across local tools further suggests false alarms.
  STEP 3 — Resolution: Favor the consistent global-real signals and clean visual evidence over conflicting local positives. Conclude no manipulation.

is_manipulated:        false
manipulation_locality: n/a
arbitrated_tools:      (none)
decision_path:
  1. Visually inspect: lighting, shadows, textures consistent; no seams.
  2. Global AI detectors (ClipSID, D3) agree image is real.
  3. Local tools conflict: RADAR strong, TruFor weak, NFA-ViT negative.
  4. Heatmaps disagree spatially; activations follow zebra stripes — false-positive prone.
  5. Integrate: global negatives + clean visuals outweigh local conflicts — pristine.
\end{lstlisting}

\noindent
The judge receives RADAR at score~1.0 and TruFor reporting a manipulation, yet correctly
identifies them as false alarms by noting that (a)~the two positive heatmaps disagree spatially,
(b)~both global detectors report real, and (c)~RADAR's activation follows zebra stripes,
a well-known false-positive surface.

\section{Two Representative OOD Cases}
\label{sec:supp-cases}

The zebra walkthrough above used a single in-distribution image to expose every prompt
and output in full.  Here we complement it with two out-of-distribution cases---one
authentic, one manipulated---presented more compactly: for each we give only the Stage-1
auditor outputs and the Stage-2 judge output, as the prompts are identical to those in
\cref{sec:supp-walkthrough}.

\subsection{Authentic: Backlit Cow at Sunset}
\label{sec:supp-pristine}
\emph{Source:} an authentic image from the CASIA~v2.0 authentic-image pool.

\begin{figure}[H]
  \centering
  \includegraphics[width=\textwidth]{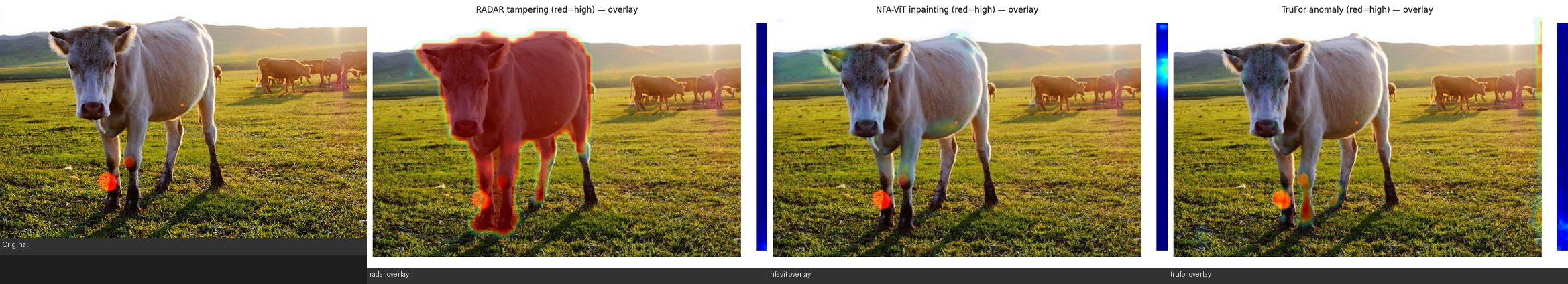}\\[4pt]
  \small{(a) Image~2 — composite overlay panel (sent to judge)}\\[8pt]
  \includegraphics[width=\textwidth]{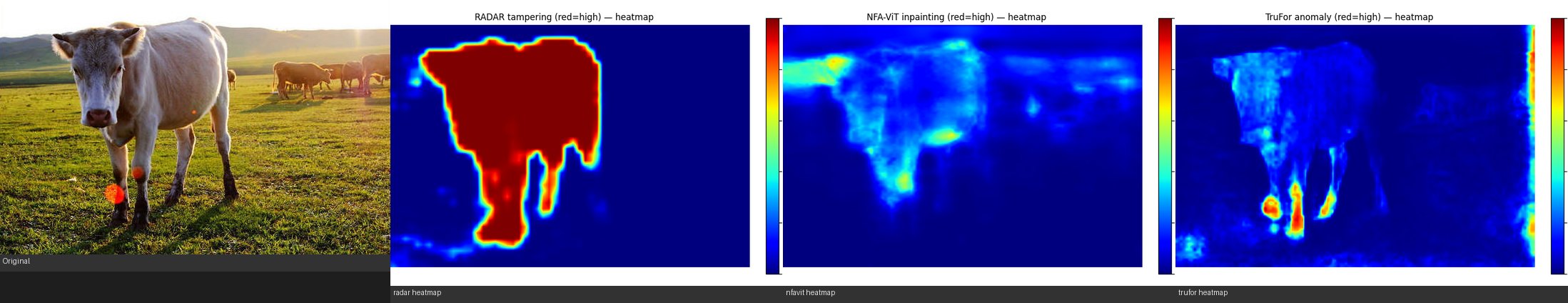}\\[4pt]
  \small{(b) Image~3 — composite raw heatmap panel (sent to judge)}
  \caption{Inputs for the authentic cow case (Case~\ref{sec:supp-pristine}).
    In each panel the top-left tile is the original; the others are RADAR, NFA-ViT, and TruFor.
    RADAR reports a manipulation at 1.0 with a whole-cow heatmap; NFA-ViT is borderline
    with a diffuse pattern; TruFor is below threshold.}
  \label{fig:supp-pristine}
\end{figure}

\FloatBarrier
\begin{lstlisting}[caption={Stage~1 auditor outputs --- authentic cow field, verbatim from
  the trace. Three detectors are invalidated (Effort and ForensicsAdapter for scope, NFA-ViT
  for a borderline / spatially incoherent signal); RADAR reports a manipulation but is alone.},
  captionpos=t,label={lst:cow-audit}]
clipsid -> valid   [is_fake=0.297 fired=False llr_score=-0.8626]
  STEP 1: The tool is designed for whole-image AI detection, which is applicable to this photograph of cattle in a field. STEP 2: The is_fake_score of 0.2968 is well below the 0.5 threshold, and the llr_score of -0.8626 indicates a clear signal towards the 'real' classification rather than being stuck at the decision boundary. STEP 3: Since the tool is in-scope and provides a coherent, interpretable signal (predicting the image is real), the output is credible evidence for downstream agents.

d3 -> valid   [is_fake=0.016 fired=False fake_score=0.0164]
  STEP 1: The image is a whole-image photograph of cattle in a field, which falls within the scope of D3's design to detect whole-image AI generation (GAN/diffusion). STEP 2: The signal is highly interpretable; the is_fake_score of 0.0164 is far from the 0.5 threshold, indicating strong confidence from the model that the image is real. STEP 3: Since the tool is in-scope and provides a clear, non-ambiguous signal, the output is credible evidence and is marked as valid.

radar -> valid   [is_fake=1.000 fired=True tamper_score=1.0000]
  STEP 1: The image contains a natural scene with animals, which is within the scope of RADAR's ability to detect localized object insertions or edits. STEP 2: The is_fake_score is 1.0, representing a maximum signal strength far from the 0.5 threshold. The heatmap (Image 2) shows a highly concentrated, coherent spatial pattern covering the primary subject (the foreground cow). This localized high-probability region corresponds directly to a specific object in the original image, providing strong spatial support for the score. The signal is interpretable and not noise-driven. STEP 3: Since the scope is valid and the signal is strong and spatially coherent, the verdict is valid.

trufor -> valid   [is_fake=0.333 fired=False integrity_score=0.3329]
  STEP 1: The image contains a natural scene with animals, which is within the scope of TruFor's ability to detect local manipulations like splicing or inpainting. STEP 2: The is_fake_score (0.3329) is below the 0.5 threshold, indicating a 'real' verdict. The heatmap shows highly localized, intense red/orange anomalies specifically on the lower legs of the foreground cow. While these are 'hot' regions, they are spatially concentrated and do not represent a global or noisy pattern. The low integrity score combined with these specific local anomalies suggests the tool is correctly identifying the image as largely authentic while noting minor local irregularities. The signal is interpretable and provides credible evidence for the downstream pipeline.

effort -> invalid   [is_fake=0.829 fired=True fake_prob=0.8291]
  STEP 1: The tool is designed to detect face-manipulation deepfakes in a single, centrally framed human face. The provided image contains cows in a field and no human faces. Therefore, the image is outside the tool's scope, making the output meaningless. Per the task instructions, a scope failure requires an 'invalid' verdict regardless of the score.

forensicsadapter -> invalid   [is_fake=0.947 fired=True fake_prob=0.9472]
  STEP 1: The tool is designed specifically for face forgery detection in single human portraits. The provided image contains cows in a field and no human faces. Therefore, the image is outside the tool's scope, making the output meaningless. Per the task instructions, a scope failure results in an 'invalid' verdict regardless of the score.

nfavit -> invalid   [is_fake=0.512 fired=True fake_score=0.5121]
  STEP 1: The tool is designed to detect inpainting (local object removal/replacement). The image contains a natural scene of cows in a field, which is within the scope of detecting local manipulations. STEP 2: The is_fake_score is 0.5121, which is extremely close to the 0.5 threshold, indicating a very weak and uncertain signal. Furthermore, the heatmap (Image 2) shows a highly diffuse, non-localized, and unstructured pattern that covers large portions of the background and the cow's body without identifying a compact, contiguous region of tampering. This lack of spatial coherence, combined with the marginal score, suggests the signal is likely noise rather than credible evidence of inpainting. Therefore, the output is deemed uninterpretable and invalid.
\end{lstlisting}

\begin{lstlisting}[caption={Real GPT-5 judge output --- authentic cow field (verbatim from trace).},
  captionpos=t,label={lst:cow-judge}]
reasoning: STEP 1 — Visual read: Backlit cow in pasture at sunset. Shadows align across animals and ground; lens-flare orbs appear in front of the cow; edges, scale, and texture look natural.
  STEP 2 — Conflicts: Whole-image AI detectors (ClipSID, D3) are silent (predict real).  Local tools disagree: RADAR fires strongly on the foreground cow; TruFor is below threshold with small hotspots on the lower legs.
  STEP 2b — Tool credibility:  RADAR: score=1.0 with a heatmap blanketing the entire foreground cow.  While attacker-plausible for splicing, such whole-object saturation is a known false-alarm pattern on salient, backlit objects; additionally, the in-front lens flare contradicts a simple post-hoc insertion.  No in-scope corroboration. TruFor: integrity 0.333 (<0.5), hotspots confined to hooves — plausible natural artifacts (shadow/flare), not a coherent tamper region.  ClipSID/D3: confidently real; they cannot confirm local edits but argue against full-image synthesis.
  STEP 3 — Resolution: Given consistent scene physics and lack of multi-tool local corroboration, downweight RADAR's lone claim. Favor the combined evidence from TruFor (under-threshold) and global detectors.  Verdict: no manipulation.

is_manipulated:        false
manipulation_locality: n/a
arbitrated_tools:      (none)
decision_path:
  1. Visually check lighting, shadows, texture, boundaries; scene appears consistent with in-camera photo.
  2. Record tool outputs: ClipSID real, D3 real, RADAR tampered, TruFor below threshold with leg hotspots.
  3. Identify conflict within local scope: RADAR vs TruFor.
  4. Assess credibility: RADAR blanket foreground-cow heatmap; no corroboration; lens flare in front contradicts splicing.
  5. Integrate evidence; weight TruFor and global detectors; conclude pristine image.
\end{lstlisting}

\noindent
Of the retained detectors, only RADAR reports a manipulation, blanketing the foreground
cow at saturation.  Lacking any in-scope corroboration and contradicted by the consistent
scene physics, the judge rejects it as a false alarm and returns \emph{authentic}.

\subsection{Manipulated: Face Swap (Celeb-DF-v2)}
\label{sec:supp-manip}
\emph{Source:} a face-swap manipulation from Celeb-DF-v2.

\begin{figure}[H]
  \centering
  \includegraphics[width=\textwidth]{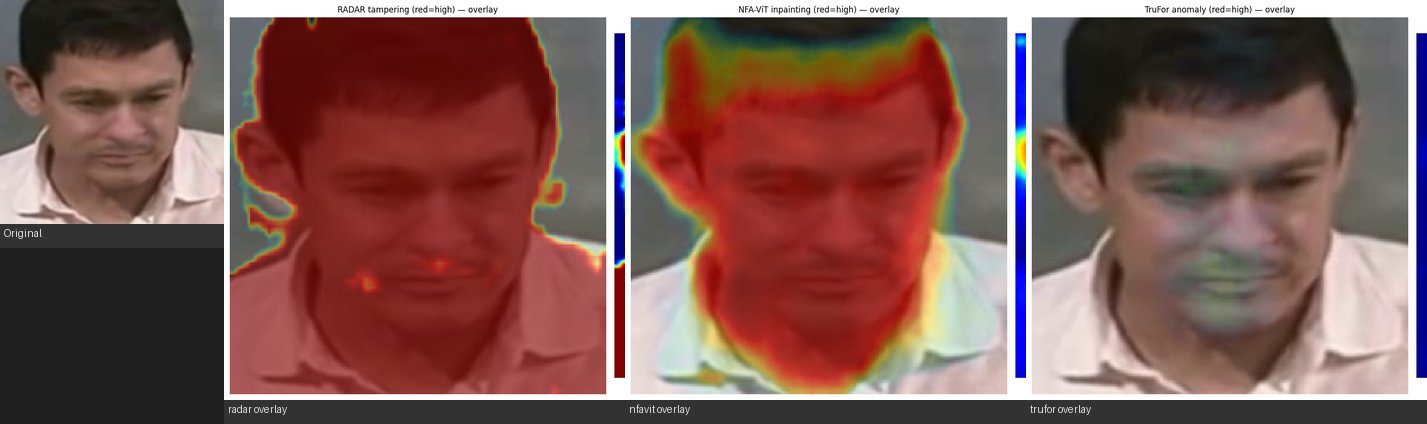}\\[4pt]
  \small{(a) Image~2 — composite overlay panel (sent to judge)}\\[8pt]
  \includegraphics[width=\textwidth]{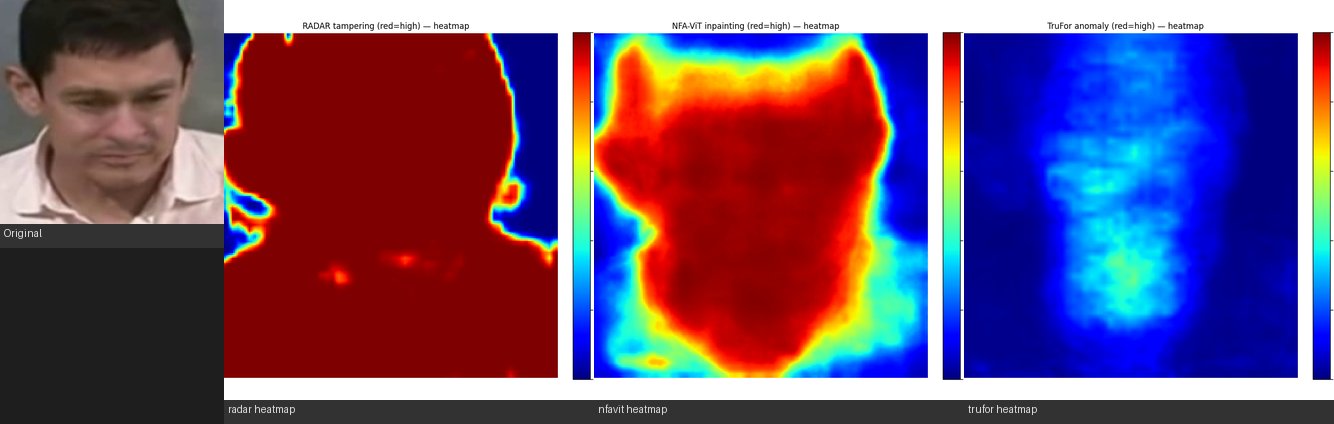}\\[4pt]
  \small{(b) Image~3 — composite raw heatmap panel (sent to judge)}
  \caption{Inputs for the face-swap case (Case~\ref{sec:supp-manip}).
    In each panel the top-left tile is the original; the others (tiles 2--4) are RADAR,
    NFA-ViT, and TruFor.
    RADAR and NFA-ViT heatmaps show saturated full-face patterns
    characteristic of false positives; TruFor is low throughout.}
  \label{fig:supp-manip}
\end{figure}

\FloatBarrier
\begin{lstlisting}[caption={Stage~1 auditor outputs --- face swap, verbatim from the
  trace. RADAR and NFA-ViT report a manipulation at saturation but are invalidated for full-face / spatially
  incoherent heatmaps; the two face-forgery specialists (Effort, ForensicsAdapter) are retained
  with high-confidence corroborating positives.},captionpos=t,label={lst:face-audit}]
clipsid -> valid   [is_fake=0.542 fired=True llr_score=0.1690]
  STEP 1: The tool is designed for whole-image AI generation detection. The image is a standard portrait, which falls within the scope of general image classification. STEP 2: The is_fake_score is 0.5421, which is slightly above the 0.5 threshold. While the signal is relatively weak (close to the decision boundary), it is a clear, interpretable value that indicates a prediction of AI-generated content. STEP 3: Since the tool is in-scope and provides a non-ambiguous signal, the output is considered credible evidence for downstream agents to weigh.

d3 -> valid   [is_fake=0.005 fired=False fake_score=0.0048]
  STEP 1: The image is a portrait of a person, which falls within the scope of a whole-image AI-generation detector (D3) designed to identify GAN or diffusion artifacts. STEP 2: The signal is highly interpretable; the is_fake_score of 0.0048 is very far from the 0.5 threshold, indicating a strong, confident prediction that the image is real. STEP 3: Since the tool is in-scope and provides a clear, non-ambiguous signal, the output is credible evidence for downstream agents.

effort -> valid   [is_fake=0.982 fired=True fake_prob=0.9819]
  STEP 1: The image contains a single human face as the dominant subject, which is the required scope for this tool. STEP 2: The is_fake_score of 0.9819 is very far from the 0.5 threshold, indicating a strong and interpretable signal. STEP 3: Since the tool is in-scope and provides a high-confidence signal, the output is credible evidence and is marked as valid.

forensicsadapter -> valid   [is_fake=1.000 fired=True fake_prob=0.9997]
  STEP 1: The image contains a single human face as the dominant subject, centrally framed in a portrait-style shot, which falls squarely within the tool's intended scope. STEP 2: The signal is extremely strong; the is_fake_score of 0.9997 is far from the 0.5 threshold, indicating high model confidence in detecting blending-boundary artifacts. STEP 3: Since the scope is met and the signal is highly interpretable and decisive, the output is credible evidence and is marked as valid.

trufor -> valid   [is_fake=0.211 fired=False integrity_score=0.2110]
  STEP 1: The tool is designed for local-manipulation detection (splicing, inpainting, etc.), which is applicable to this portrait image. STEP 2: The is_fake_score is 0.211, well below the 0.5 threshold, indicating a 'real' prediction. The heatmap shows low-magnitude, diffuse activations primarily around facial features (eyes, nose, mouth) rather than concentrated, high-intensity red regions. This pattern is consistent with low-level noise or texture-based activations common in real images rather than a coherent manipulation signal. Since the signal is interpretable and the score is far from the decision boundary, the tool's negative finding is credible evidence.

nfavit -> invalid   [is_fake=0.992 fired=True fake_score=0.9916]
  STEP 1: The tool is designed to detect localized inpainting (removing/replacing objects). While the image contains a face, the tool's output in STEP 2 shows a massive, non-localized, and saturated heatmap covering almost the entire head and neck area. This 'full-face' activation pattern is diagnostically inconsistent with the tool's purpose of identifying specific inpainting regions; it behaves more like a global texture or compression detector rather than a localized inpainting detector. Because the signal is spatially incoherent and lacks a compact, nameable target region, it is considered a false positive/uninterpretable signal, leading to an 'invalid' verdict.

radar -> invalid   [is_fake=1.000 fired=True tamper_score=1.0000]
  STEP 1: The tool is designed to detect localized diffusion-based manipulations (object insertion/removal). The input image is a portrait of a person, which is within the scope of detecting localized edits. STEP 2: The is_fake_score is 1.0, indicating a very strong signal. However, the heatmap (Image 2) shows a massive, saturated red region covering almost the entire subject, including the face and hair, with high-intensity edges around the silhouette. This uniform, high-magnitude pattern is characteristic of a low-information/saturated signal rather than a localized edit. Instead of identifying a specific manipulated object or region, the tool is essentially flagging the entire subject as anomalous, which suggests the signal is not interpretable as a specific localized tampering event. Therefore, the signal is not credible.
\end{lstlisting}

\begin{lstlisting}[caption={Real GPT-5 judge output --- face swap (verbatim from trace).},
  captionpos=t,label={lst:face-judge}]
reasoning: STEP 1 — Visual: Single centrally framed face, low resolution/compressed.  Lighting/background consistent; facial boundary (cheeks/jaw/forehead) looks unusually soft, compatible with blending.
  STEP 2 — Conflicts: (a) Whole-image AI-gen: ClipSID fired (0.542) vs D3 strongly real (0.005). (b) Local manipulation: face-forgery detectors Effort (0.982) and ForensicsAdapter (0.9997) fired, while TruFor predicted real (0.211).
  STEP 2b — Reliability: D3 is far below boundary — confident real; ClipSID is only slightly positive — likely false alarm.  Effort and ForensicsAdapter are in-scope (single dominant face) and give saturated positives — credible and mutually corroborating. TruFor's heatmap in the panels is diffuse/low around normal facial textures, a known weak pattern for face-swap detection at low resolution — down-weighted.
  STEP 3 — Resolution: Prefer D3 over ClipSID — not globally AI-generated.  Prioritize the corroborated, high-confidence face-forgery signals (ForensicsAdapter, Effort) over TruFor's weak negative.  Integrated with visual cues, the most plausible explanation is a local face manipulation (swap/reenactment) centered on the face.

is_manipulated:        true
manipulation_locality: local
manipulation_region:   face region (central portrait), especially along the facial boundary around cheeks, jawline, and forehead
arbitrated_tools:      forensicsadapter, effort
decision_path:
  1. Visual review: single face; low-res; soft facial boundary suggests blending.
  2. Conflict group: global AI-gen detectors disagree (ClipSID+, D3-).
  3. Conflict group: face-forgery tools fired; TruFor negative.
  4. Reliability: D3 strong real; ClipSID marginal — discount global AI-gen.
  5. Reliability: ForensicsAdapter and Effort high-confidence, in-scope, corroborating.
  6. Heatmaps: TruFor diffuse/low — likely miss; downweight.
  7. Integrate — local face manipulation centered on face.
\end{lstlisting}

\noindent
This is the only manipulated case.  The auditor retains the two in-scope face specialists
(Effort, ForensicsAdapter), which report the swap with high confidence, and excludes RADAR
and NFA-ViT, whose saturated full-face heatmaps are out-of-scope false positives.  The judge
arbitrates to the face specialists and returns a local face manipulation---the correct
verdict and the correct decisive detectors.

\section{Variant Input Differences}
\label{sec:supp-variants}

The output schema is identical across all variants; only their inputs differ.
Algorithmic fusion uses no MLLM and is described in the main paper.

\subsection{Unguided Judge}

The unguided-judge variant is identical to the full pipeline except that three sections are
removed from the judge's REASONING PROTOCOL---this is the \emph{prompting} axis of the main
paper's dissection (the judge no longer receives explicit false-positive guidance).
\Cref{lst:ablated-proto} shows the full pipeline's REASONING PROTOCOL with the removed
blocks in \textcolor{dropcol}{\itshape gray italic}.
After STEP~1 visual inspection the judge moves directly to a simplified one-step
integration (the ablated replacement is shown in black after the gray blocks).

\begin{lstlisting}[caption={Judge REASONING PROTOCOL.
  \textcolor{dropcol}{\itshape Gray italic}: blocks present in the full pipeline
  but removed in the unguided-judge variant (the judge prompt stops after STEP~1 and
  jumps to the simplified STEP~2 shown in black).},
  label={lst:ablated-proto},captionpos=t]
STEP 1 -- Independent semantic visual inspection of Image 1
   [... identical in both variants: describe lighting, texture, perspective, boundaries, AI-generation cues ...]

|STEP 2 -- Identify potential conflicts among detectors
   Every surviving tool has two attributes -- SCOPE (from the tool card's scope_in) and OUTPUT (fired / silent) -- that determine whether its verdict is consistent with another tool's. The governing principle: at most ONE scope can be the true scope of a manipulation.  Walk through the surviving tools, compare their (scope, output) pairs, and list the conflict groups you find. Do NOT attempt resolution yet -- that is STEP 3.

STEP 2b -- Reason about per-tool reliability on THIS image
   Each surviving tool is a statistical classifier whose accuracy is below 100 %.  The reliability lens for this image:
     - Scope vs image content: does Image 1 plausibly contain the manipulation the tool detects?
     - Score magnitude relative to the decision boundary.
     - Heatmap shape (for heatmap tools): compact on an attacker-plausible region vs scattered on natural-texture content where spurious activations are well-documented?
     - Cross-tool corroboration within the same scope.

STEP 3 -- Resolve conflicts and integrate into the final verdict
   For each conflict listed in STEP 2, decide which direction survives using: tool cards' scope_in vs Image 1 content; composite heatmap panels; auditor reasoning from the audit table. Integrate surviving evidence with STEP 1 visual observations.|

STEP 2 -- Integrate tool evidence into the final verdict
   Using your STEP 1 visual observations, the tool cards, the audit table, and the composite heatmap panels, decide whether the image has been manipulated.  When surviving tools and visual observations point in different directions, you decide how to weigh them given the inputs available to you.
\end{lstlisting}

\noindent
The per-case inputs (retained tool cards, audit evidence) are \emph{identical}
to the full pipeline.

\subsection{Single-Agent}

The seven per-detector auditor calls are removed entirely.
A single judge agent receives \emph{all seven} detectors' raw scores and tool cards
(no pre-filtering) and is tasked with doing both jobs in one pass.

\begin{lstlisting}[caption={Judge prompt for the single-agent variant.
  The STEP~0 self-audit block replaces the separate auditor stage.},
  captionpos=t,label={lst:noaudit-proto}]
You are the final judge agent.  In this configuration there is NO separate per-tool auditor -- you do BOTH jobs in one pass: first assess each detector's credibility on this image yourself, then integrate the credible ones into a single forensic verdict.

-------------- TASK --------------
[... identical structure but receives ALL seven detectors' raw evidence (no pre-filtering) and all seven tool cards (not just the surviving ones).]

-------------- REASONING PROTOCOL --------------

STEP 0 -- Per-tool credibility  (you are the auditor here)
   For EACH detector, decide whether its output is credible:
     - SCOPE: does Image 1 contain the content this detector was designed for?  Out-of-scope score is meaningless -- set aside.
     - SIGNAL: if scope holds, is the score interpretable (not stuck near the 0.5 boundary)?  For heatmap detectors (Images 2 and 3): does the spatial pattern land on an attacker-plausible region, or scatter across natural-texture content where spurious activations are well-documented?
     - Cross-tool corroboration within a shared scope: a lone positive among silent same-scope peers is more likely a false alarm.
   Name each detector as kept or set aside, with reason. Carry the credible detectors forward.

STEP 1 -- Independent visual inspection of Image 1
   [... same instructions as the full pipeline ...]

STEP 2 -- Identify conflicts among credible detectors
   [... same single-scope principle applied to STEP 0 survivors ...]

STEP 3 -- Resolve conflicts and integrate into final verdict
   [... same as the full pipeline ...]

[OUTPUT FORMAT -- identical to the full pipeline.]
\end{lstlisting}

\noindent
The full pipeline runs seven parallel MLLM auditor calls plus one judge call ($\approx$8
calls total); the single-agent variant collapses this to a \textbf{single} MLLM call.
The judge prompt grows correspondingly (all seven tool cards, no pre-assigned verdicts),
and STEP~0 adds a self-audit pass before the conflict-resolution steps.

\end{document}